\documentclass[11pt]{article}
\usepackage[preprint]{acl}
\usepackage{times}
\usepackage{latexsym}

\usepackage[T1]{fontenc}
\usepackage[utf8]{inputenc}
\usepackage{microtype}
\usepackage{inconsolata}
\usepackage{graphicx}

\usepackage{tipa}

\usepackage{times}
\usepackage{latexsym}
\usepackage[T1]{fontenc}
\usepackage[utf8]{inputenc}
\usepackage{microtype}
\usepackage{inconsolata}
\usepackage{graphicx}

\usepackage{booktabs}      %
\usepackage{amsmath}
\usepackage[table]{xcolor} %
\usepackage{subcaption}    %
\usepackage{epigraph}

\renewcommand{\eqref}[1]{Equation~\ref{eq:#1}}
\newcommand{\secref}[1]{Section~\ref{sec:#1}}
\newcommand{\secstworef}[2]{Sections~\ref{sec:#1} and~\ref{sec:#2}}

\newcommand{\appref}[1]{Appendix~\ref{app:#1}}
\newcommand{\appstworef}[2]{Appendices~\ref{app:#1} and~\ref{app:#2}}
\newcommand{\figref}[1]{Figure~\ref{fig:#1}}

\newcommand{\promptref}[1]{Prompt~\ref{pro:#1}}
\newcommand{\promptstworef}[2]{Prompts~\ref{pro:#1} and~\ref{pro:#2}}

\newcommand{\tabref}[1]{Table~\ref{tab:#1}}

\DeclareRobustCommand{\DE}[3]{#2}

\usepackage[colorinlistoftodos,prependcaption,textsize=tiny,textwidth=14mm]{todonotes}
\makeatletter
\define@key{todonotes}{AW}[]{%
    \setkeys{todonotes}{author=AW,color=blue!20}}%
\makeatother   
\makeatletter
\define@key{todonotes}{WL}[]{%
    \setkeys{todonotes}{author=WL,color=red!20}}%
\makeatother   

\title{The Hrunting of AI: Where and How to Improve English Dialectal Fairness}

\author{
    \textbf{Wei Li\textsuperscript{\rm 1}}\thanks{Work during an internship at Microsoft.} and
    \textbf{Adrian de Wynter\textsuperscript{\rm 2, 3}}
\\
\\
    \textsuperscript{\rm 1}Boston College,
    \textsuperscript{\rm 2}Microsoft,
    \textsuperscript{\rm 3}The University of York
\\
  \small{
    \textbf{Correspondence: }\href{mailto:adewynter@microsoft.com}{adewynter@microsoft.com}
  }
}

\begin{document}

\maketitle

\begin{abstract}
It is known that large language models (LLMs) underperform in English dialects, and that improving them is difficult due to data scarcity. 
In this work we investigate how quality and availability impact the feasibility of improving LLMs in this context. 
For this, we evaluate three rarely-studied English dialects (Yorkshire, Geordie, and Cornish), plus African-American Vernacular English, and West Frisian as control. 
We find that human-human agreement when determining LLM generation quality directly impacts LLM-as-a-judge performance. 
That is, LLM-human agreement mimics the human-human agreement pattern, and so do metrics such as accuracy. 
It is an issue because LLM-human agreement measures an LLM's alignment with the human \textit{consensus}; and hence raises questions about the feasibility of improving LLM performance in locales where low populations induce low agreement. 
We also note that fine-tuning does not eradicate, and might amplify, this pattern in English dialects. 
But also find encouraging signals, such as some LLMs' ability to generate high-quality data, thus enabling scalability. 
We argue that data must be carefully evaluated to ensure fair and inclusive LLM improvement; and, in the presence of scarcity, new tools are needed to handle the pattern found. 

\end{abstract}

\epigraph{
\begin{minipage}[t]{.47\textwidth}
Ǣghwæþres sceal\\
scearp scyldwiga gesc\=ad witan,\\
worda ond worca, s\=e þe w\=el þenceð.
\end{minipage}
\hfill%
\begin{minipage}[t]{0.47\textwidth}
A sharp shield-warrior \\
must be a judge of both things, \\ 
words and deeds, if he would think well.
\end{minipage}
}{
\textit{Unknown}, Beowulf IV 285-289 (t. by R. Liuzza)
}

\section{Introduction}\label{sec:intro}

\textit{Beowulf} is the oldest surviving poem in English, dating to circa 975 C.E. 
It is written in Late West Saxon and dialects of Old English, the precursors to Modern English and close relatives of contemporary West Frisian. 
The poem provides invaluable insights on the history, customs, and values of ancient Anglo-Saxon culture. 
Crucially, it illustrates the evolution of the English language, including its variations and dialectal origins. 

Today, Modern English is spoken by an estimated 370M people as L1, and 978M as L2 \cite{lewis2009ethnologue}, and dominates online communication \cite{joshi-etal-2020-state}. 
As a pluricentric language, it has never been strictly standardised. 
It encompasses many dialects and variations, such as Yorkshire English and African-American Vernacular English (AAVE), each reflecting distinct identities, histories, and cultural identities.

Even though large language models (LLMs) are excellent at complying with user requests, this promise has shown to hold mostly for mainstream English, and not for its dialects \cite{pan-etal-2025-analyzing,ryan-etal-2024-unintended}. This in turn risks marginalising those perceived as not speaking the `correct' or `educated' version of the language \cite{milroy}.\footnote{Hrunting was a sword which `had never failed any man who grasped it in his hands' (XX L1460-1461). However, `the edge failed the man [Beowulf] in his need' (XX 1524-1525), hence the title of this work.} 
However, addressing this disparity is challenging due to the scarcity of high-quality, human-produced data, plus the absence of validated synthetic corpora.%

\begin{figure}[ht]
    \centering
    \includegraphics[width=\linewidth]{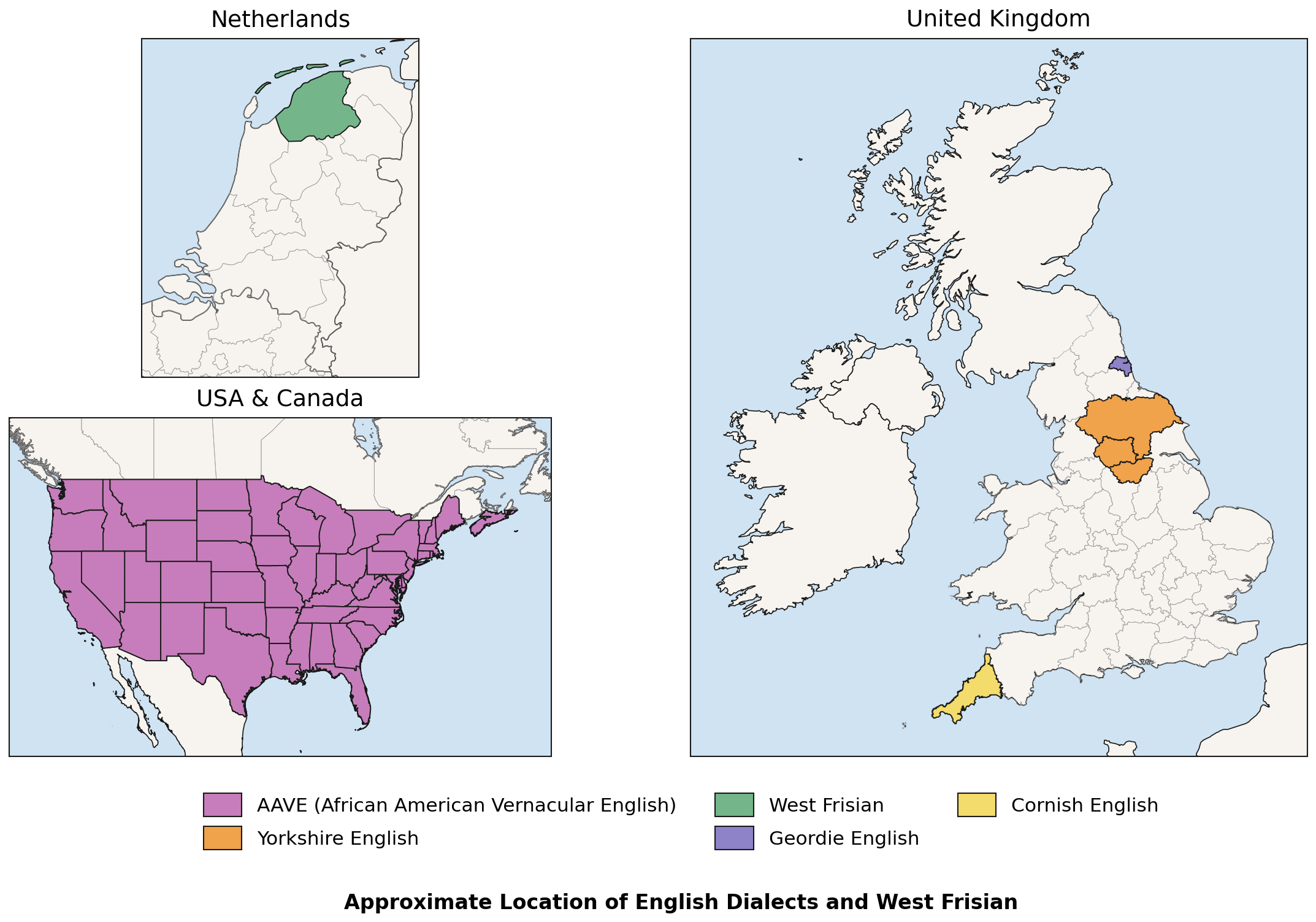}
    \caption{Geographical distribution for the English dialects used in this work and our control, West Frisian.}
    \label{fig:dialectsused}
\end{figure}

In this work we explore to which extent data quality impacts the feasibility of the development of effective and inclusive systems in English dialects. 
We perform a full exploration of the end-to-end standard data pipeline: generation, human evaluation, machine (LLM) evaluation, and fine-tuning. 
Our evaluation is on four English dialects (\figref{dialectsused}) and West Frisian, a low-resource close relative of English, acting as a control \cite{lewis2009ethnologue}.\footnote{See \appref{comprehensivedialects} for further discussion on this.} 
Its linguistic features make it an ideal baseline to obtain insights into the complexity of tackling this problem with respect to English dialects. 

\subsection{Findings}

We find evidence that human-human annotation agreement on generation quality judgements directly impacts an LLM's ability to perform the same judgement. 
That is, the pattern of agreement distribution is mimicked by the human-LLM agreement. It is also mimicked by downstream performance metrics (e.g., F$_1$) in both in-context learning (ICL) and fine-tuned scenarios. 

This is an issue because human-LLM agreement measures an LLMs concordance with the \textit{human consensus}, and suggests that alignment methods are not sufficiently effective for English dialects. 
This also matters because low-resource, non-standardised dialects often have few speakers available to ensure high-agreement annotations. Indeed, we note that the pattern could stem from resourceness and linguistic features intrinsic to English dialects. 
All of this impedes the feasibility of building and specialising LLMs for these \textit{dialects}, as the results do not extend to West Frisian. 

That said, we also find other kinds of results: 

\begin{enumerate}
    \item \textbf{Positive results}, such as signals of existing improvement for some LLMs with respect to previous findings. 
    For example, we find that over 92\% of LLM responses in some locales (AAVE, Yorkshire English) are considered high quality by native speakers. 
    \item \textbf{Encouraging results}, such as the viability of effective LLMs-as-judges in some locales, thus enabling a feasible, scalable path for improving LLMs. 
    \item \textbf{Negative results}, where the development of linguistics-informed tools is likely required to provide a marked betterment of LLMs. 
    For example, we note that low human-human agreement leads to synthetic data and direct preference optimisation (DPO; \citealt{10.5555/3666122.3668460}) approaches to be ineffective at, and sometimes detrimental to, LLM fine-tuning. 
\end{enumerate}

In addition to related work (\appref{background}) we also provide a short-but-comprehensive survey of dialect studies between 2024 and 2026 (\appref{litreview}). 
Our findings inform the feasibility of creation and assessment of dialect-specific, task-oriented corpora; but also highlight the difficulty on ensuring full and culturally attuned development of LLMs.\footnote{Data and code is available at \url{https://github.com/adewynter/Beowulf}.}

\section{Methods}

\subsection{Languages and Dialects Covered}
We cover the following English dialects: African American Vernacular (en-US-AAVE), Cornish (West Country; en-UK-Wes), Geordie (en-UK-Geo), and Yorkshire (en-UK-York) English. 
We chose these due to their morphosyntactic characteristics, historical development, and literature coverage. 
For control purposes, we also consider West Frisian (fy-NL), which we selected due to its linguistic closeness to English and lack of studies around it. 
Although it is meant to act as control, we fully consider it in our evaluations. 
More details plus considerations on the definition of a dialect are in \appstworef{comprehensivedialects}{litreview}. 

\subsection{Approach}\label{sec:approach}
We evaluate the impact of data scarcity with respect to performance in the `standard pipeline' to build data for fine-tuning an LLM: (1) creating task-specific data (TSD), which involves developing synthetic locale-specific corpora (prompts and outputs) to specialise a model, ideally with human supervision; and (2) evaluating automated TSD, which requires the development of automated methods (usually LLMs-as-judges) to enable scalability. 

\subsection{Rubric and Annotation Metrics}\label{sec:annotation}

We measure TSD quality following a rubric-based approach. 
This has the advantage of ensuring transparency in the annotation process, and provide a further comparison point in our experiments. 
Our rubric has the following binary criteria, evaluating specifically the LLM's output:

\paragraph{c1:} It must be in the locale specified.
\paragraph{c2:} It must be culturally (e.g., using the right measurement units) and argumentatively (i.e., it should make sense) accurate, and contain an explanation.
\paragraph{c3:} It must be correct and not contain syntax errors in the case of code.
\paragraph{c4:} It must be grammatically correct and/or free of syntax errors.
\paragraph{c5:} It must not be cut off.
\paragraph{c6:} The model must follow the prompt exactly and completely. It cannot refuse to respond and should continue writing if the prompt lacks an explicit instruction.\\

The final label is label=1 if none of the criteria are zero, otherwise label=0. 
These are criteria prescribe a highly specific behaviour, and are meant to evaluate an LLM's ability to improve in an arbitrary context. 
This is because since LLMs could be developed for well-defined downstream applications.
We test other variations in \secref{ablationrubric}. 
For the full rubric, see \appref{rubrics}.

In addition to accuracy and F$_1$, we report various measures of statistical significance, primarily with Gwet's AC1 for agreement, and McNemar' test when comparing LLM outputs. 
AC1 is particularly useful since it is not as sensitive to skewed data as Cohen's $\kappa$ \cite{CICCHETTI1990551}, and it is designed for multiple raters. 
It may be interpreted the same way as Cohen's $\kappa$, with values above 0.8 indicating near-perfect agreement and below 0.2 near random \cite{gwet,fleiss}.

\subsection{TSD Collection}\label{sec:datacollection}

We built the TSD corpora by transcreating 1,000 randomly-sampled task-specific prompts from existing en-US-corpora:

\noindent\textbf{WildChat:} a general-conversation corpus of user-submitted prompts to GPT models \cite{zhao2024wildchat}.\\
\textbf{GSM8K:} grade school maths word problems \cite{zhang2024careful}.\\
\textbf{OpenOrca:} a general-purpose multi-task dataset \cite{mukherjee2023orca}. \\
\textbf{OpenCode:} a reasoning-based synthetic dataset for coding \cite{ahmad2025opencodereasoning}. \\
\textbf{Stanford Human Preferences (SHP):} human preference annotations on prompt responses. We focus on \textbf{step-by-step planning} (cooking and baking) \cite{pmlr-v162-ethayarajh22a}.\\

Every per-locale corpus contains the same prompts. Their corresponding outputs were created by prompting the LLM with the task-specific prompt, and included the rubric in the system prompt. 
See \appref{prompts} for the prompts used and \appref{detailedmethods} for licencing and sample entries. %

\subsubsection{Human Annotation}\label{sec:datastats}
All data was annotated per-criterion by three native speaker annotators, following the rubric from \appref{rubrics}. 
The average AC1 between annotators was 0.75, with AAVE and Yorkshire having the highest agreements (0.98 and 0.96), followed by Geordie, Frisian, and Cornish (0.67, 0.61, 0.51). 
These numbers all indicate near-perfect or moderate agreement. 
The aggregate (final) label was picked as a per-criterion majority vote. 
See \secref{ethics} for further details on the annotation process.

\subsection{LLMs Used}
We used the following models: GPT-5-mini \cite{GPT5}, GPT-4.1 \cite{GPT41}, Claude Opus 4.1 \cite{ClaudeSonnet}, Phi-4 \cite{abdin2024phi4technicalreport}, and Qwen3 8B \cite{yang2025qwen3technicalreport}. 
We evaluated all LLMs but Phi-4 in both TSD generation and evaluation. 
Due to budget considerations, we only employed Phi-4 in the latter. 
We chose Qwen3 for our fine-tuned judge. 
Call parameters and prompts are in \appstworef{llmmethods}{prompts}.

\section{Results}\label{sec:results}

\subsection{TSD Generation}\label{sec:rq1results}

To determine quality of the generations, we measured the proportion of label 1 aggregates as judged by the human annotators. 
The results showed a disparity per-model and locale. The average generation quality was high in AAVE and Yorkshire (95.7 and 97.5\% for all models but Qwen3), but not in Cornish (40.4\%; without Qwen3), Geordie (r. 61.2\%), or West Frisian (r. 27.7\%). Qwen3 had the lowest performance (51.2\%), followed by GPT-5 (60.2\%), GPT-4.1 (70.3\%), and Opus-4.1 (72.5\%).

\subsubsection{Error Analysis}

We observed that the lowest-rated task was coding in all locales but West Frisian, where it was a tie between it and SHP. Coding was particularly difficult for all models, with Qwen3 and GPT-5 often scoring last and second-last on these (e.g., 7.7\% versus 15.1\% in Cornish). 

The most frequent comment by annotators in the lowest-performing locales (Cornish and West Frisian) was that LLMs generated output in the wrong locale, e.g. Cornish (the language; 212 annotator responses) or Dutch (r. 216). 
In West Frisian annotators often mentioned that LLMs had spelling (r. 215) and/or grammatical (r. 1023) errors in addition to misusing terms. 
Note that the prompts explicitly stated the locale. 
Geordie had mostly low quality outputs (inaccurate and wrong). 
In contrast, the outputs from AAVE, when considered incorrect, were due to them not being in the locale (86 instances; 53 were in en-US-std). For Yorkshire, it was poor prompt compliance. 
English dialects, particularly Cornish and Geordie, often found Scottish or Northern words in inadequate contexts. 
See \tabref{label1aggr} and \appref{hqbysource} for further details.

\begin{table}[]
    \centering
    \begin{tabular}{lp{0.13\linewidth}p{0.13\linewidth}p{0.13\linewidth}p{0.13\linewidth}}\toprule
    \textbf{Locale} & \textbf{Opus 4.1} & \textbf{GPT-5} & \textbf{GPT-4.1} & \textbf{Qwen3 8B} \\ \midrule
    \textit{AAVE     } & 96.4 & 95.6 & 95.2 & 79.8 \\
    \textit{Cornish  } & 58.3 & 29.3 & 55.1 & 18.7 \\
    \textit{Geordie  } & 71.0 & 53.3 & 64.4 & 56.0 \\
    \textit{Yorkshire} & 97.6 & 96.8 & 98.0 & 94.2 \\
    \textit{Frisian  } & 39.0 & 25.9 & 38.6 & 7.2 \\ \bottomrule
    \end{tabular}
    \caption{Percentage of outputs considered high quality (label 1) by human annotators. Most LLMs had good performances in Yorkshire and AAVE, but struggled with Cornish, Geordie, and West Frisian. 
    The latter had the lowest average performance, particularly for Qwen3.}
    \label{tab:label1aggr}
\end{table}

\subsection{Evaluating TSD}\label{sec:rq2results}

We compared various prompting and algorithmic strategies to determine the ability of LLMs to evaluate TSD quality. These are (1) ICL strategies; (2) fine-tuned models; and variations by decomposing them by criteria. 

\subsubsection{ICL Approaches}\label{sec:icl}
We tested five main strategies:

\noindent\textbf{Strategy 1:} Output the final label. \\
\textbf{Strategy 2:} Output the label and a rationale.\footnote{This is known to increase robustness in some scenarios \cite{Brahman_Shwartz_Rudinger_Choi_2021}. }\\
\textbf{Strategy 3:} Score each criterion, and then return the final label and rationale. \\
\textbf{Strategy 4:} Score one criterion per call. Then aggregate the responses programmatically.\\
\textbf{Strategy 5:} Same as strategy 4, but include a rationale in every response. 

All strategies were given the full rubric (or criterion specification), and were tested with zero, five and twenty exemplars. 
We observed no significant differences or patterns in all prompt strategies with respect to shots, with average accuracy standard deviations of $\pm 4.6$ per model per locale. 
The positive contributions were from five-shot; hence we focus on these results. 
Strategies 1-3 were significantly (+5\% accuracy, +11.4 F$_1$ average) more effective than aggregating per criterion (strategies 4 and 5), especially when requesting \textit{only} the label (strategy 1). 
We also did not observe noticeable changes in performance on strategies 1-3, with average standard deviations of $\pm0.5$ (accuracy), $\pm0.74$ (F$_1$), and $\pm0.11$ (AC1) per model per locale. An ordinary least squares (OLS) fit over per-shot accuracy indicated barely-changing performance (-0.1, 0.6, and 0.8 for strategies 1-3). 

\begin{figure}
    \centering
    \includegraphics[width=0.95\linewidth]{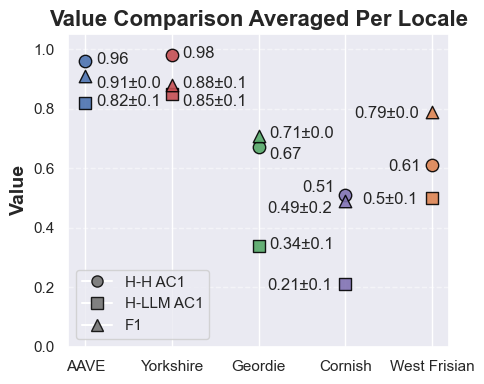}
    \caption{Average AC1 between humans (H-H) and humans and LLMs (H-LLM), compared with the LLMs' averaged F$_1$, with the same dataset and (best-performing) ICL prompt. 
    H-H roughly aligns with H-LLM, and with the LLMs' F$_1$. H-LLM, in theory, should be near 1.0, even if H-H is not, to indicate agreement with the human consensus. 
}
    \label{fig:agreementsaggregate}
\end{figure}

LLMs had varying performances and agreements per locale. 
In strategy 1, AAVE and Yorkshire had good F$_1$, accuracy, and AC1, followed by West Frisian and Geordie, and trailed by Cornish. The human-LLM agreement and F$_1$ followed the same pattern as the human-human agreement (\figref{agreementsaggregate}). The full results for strategy 1 are in \tabref{baselineresults}. 
Further details are in \secref{ablation} and \appref{baselinejudgesdetails}. 

\begin{table*}[]
    \centering
    \begin{tabular}{lcccccccccc|cc}
\toprule
\textbf{Locale} & \multicolumn{2}{c}{\textbf{Opus 4.1}} & \multicolumn{2}{c}{\textbf{GPT-5}} & \multicolumn{2}{c}{\textbf{Qwen3 8B}} & \multicolumn{2}{c}{\textbf{GPT-4.1}} &  \multicolumn{2}{c}{\textbf{Phi-4}} &  \multicolumn{2}{|c}{\textbf{Average}} \\
            &  Acc & F$_1$ & Acc & F$_1$ & Acc & F$_1$ & Acc & F$_1$ & Acc & F$_1$ & Acc & F$_1$ \\ \midrule
\textit{AAVE} & \cellcolor{red!10}74.7 & \cellcolor{red!10}84.6 & 86.0 & 92.0 & 85.0 & 91.5 & \cellcolor{blue!10}90.9 & \cellcolor{blue!10}95.1 & 81.9 & 89.8 & 83.7 & 90.6 \\
\textit{Yorkshire} & 71.3 & 82.8 & 87.2 & 93.0 & 86.6 & 92.7 & \cellcolor{blue!10}91.0 & \cellcolor{blue!10}95.2 & \cellcolor{red!10}62.1 & \cellcolor{red!10}75.9 & 79.6 & 87.9 \\
\textit{Geordie} & \cellcolor{red!10}57.3 & \cellcolor{red!10}64.4 & 59.7 & 71.8 & 60.1 & 72.8 & \cellcolor{blue!10}61.7 & \cellcolor{blue!10}74.4 & 59.7 & 70.5 & 59.7 & 70.8 \\
\textit{Cornish} & 60.5 & 60.8 & \cellcolor{red!10}54.2 & \cellcolor{red!10}0.0 & 54.2 & 62.2 & 54.7 & 60.2 & \cellcolor{blue!10}54.0 & \cellcolor{blue!10}62.7 & 55.5 & 49.2 \\
\textit{West Frisian} & \cellcolor{blue!10}82.7 & \cellcolor{blue!10}83.7 & 79.6 & 78.5 & \cellcolor{red!10}68.0 & \cellcolor{red!10}72.0 & 78.5 & 81.3 & 74.4 & 78.4 & 76.6 & 78.8 \\
\bottomrule
    \end{tabular}
    \caption{Accuracy and F$_1$ for strategy 1. The best-performing (by F$_1$) LLM is in blue; the lowest in red. All LLMs had good performances in AAVE and Yorkshire, but lower in West Frisian (except Opus 4.1), Geordie, and Cornish. 
    }
    \label{tab:baselineresults}
\end{table*} 

\subsubsection{Fine-tuned Results}

The second TSD evaluation involved fine-tuning Qwen3 8B on strategies 1 and 4. 
We evaluated three methods: supervised fine-tuning (SFT), SFT with synthetic data expansion (transcreated and labelled by the best-performing LLM per locale), and DPO. 
Synthetic data expansion is a natural choice when datasets are scarce, and DPO is meant to be more data efficient and designed to align LLMs with human preferences. 
For this we split the human dataset at a 80/20 train/test uniformly-at-random selection by indices--thus we used exact same test set for all locales. Due to the test sets' relatively small size ($n=203$), we place more importance on measures of statistical significance and AC1 than accuracy or F$_1$. 

Human-based SFT improved LLMs in both strategies, with the largest increase in AC1 (Cornish) coming from strategy 1 at +0.25 (+14.8 accuracy, +3.3 F$_1$, $p$ < 0.001). This signified a positive change of +0.31 AC1 (+14.7\% accuracy, +16.2 F$_1$) in coding \tabref{ft_vs_baseline}. 
Similar numbers were obtained for strategy 4, albeit at a smaller F$_1$ change (+3.8). 
A closer analysis showed that only coding improvements were consistently statistically significant in most locales. We elaborate on this in \secref{ablationpreds}. 

DPO and synthetic data SFT underperformed with respect to human-based SFT. For DPO, due to budgetary reasons, we only tested strategy 1 and Cornish only. 
All metrics dropped, with -17.3\% accuracy, -3.6 F$_1$, and -0.27 AC1, at $p$ < 0.001. 
For synthetic data SFT, we tested Yorkshire, West Frisian, and Cornish. 
We observed more variable performance in this subset, with Cornish having drops of -9.4\%, -3.4, and -0.10; but Yorkshire and West Frisian having increases in all metrics, particularly a +0.10 and +0.09 AC1, respectively, at $p$ < 0.001. 
No approach beat human-based SFT. 
Further details are in \appref{othertrainingstrats}.

\begin{table*}
\centering
\begin{tabular}{l  c c c c c c c c c  c c c }
\toprule
 & \multicolumn{3}{c}{\textbf{Strategy 1}} & \multicolumn{3}{c}{\textbf{Strategy 1 (FT)}} & \multicolumn{3}{c}{\textbf{Strategy 4}} & \multicolumn{3}{c}{\textbf{Strategy 4 (FT)}}  \\
             &  Acc & F$_1$ & AC1 & Acc & F$_1$ & AC1 & Acc & F$_1$ & AC1 & Acc & F$_1$ & AC1  \\ \midrule
\textit{AAVE        } & 87.2 & 92.9 & 0.85 & \cellcolor{blue!10}91.6 & \cellcolor{blue!10}95.6 & \cellcolor{blue!10}0.91 & 85.2 & 91.8 & 0.82 & \cellcolor{blue!10}91.1 & \cellcolor{blue!10}95.4 & \cellcolor{blue!10}0.90  \\
\textit{Cornish     } & 55.2 & 63.7 & 0.15 & \cellcolor{blue!10}70.0 & \cellcolor{blue!10}67.0 & \cellcolor{blue!10}0.40 & 55.7 & 64.3 & 0.16 & \cellcolor{blue!4}63.1 & \cellcolor{blue!4}67.8 & \cellcolor{blue!4}0.28  \\
\textit{Geordie     } & 64.0 & 75.4 & 0.41 & 62.6 & 75.6 & 0.42 & 61.6 & 74.2 & 0.34 & 60.6 & 75.2 & 0.41 \\
\textit{Yorkshire   } & 84.2 & 91.2 & 0.81 & \cellcolor{blue!10}96.0 & \cellcolor{blue!10}98.0 & \cellcolor{blue!10}0.96 & 85.6 & 92.1 & 0.83 & \cellcolor{blue!10}95.5 & \cellcolor{blue!10}97.7 & \cellcolor{blue!10}0.95 \\
\textit{West Frisian} & 71.9 & 74.7 & 0.45 & \cellcolor{blue!10}76.8 & \cellcolor{blue!10}75.1 & \cellcolor{blue!10}0.54 & 73.6 & 74.5 & 0.47 & \cellcolor{blue!10}80.2 & \cellcolor{blue!10}83.0 & \cellcolor{blue!10}0.61\\
\bottomrule
\end{tabular}
\caption{Accuracy, F$_1$ and AC1 for strategies 1 and 4 (fine-tuned and ICL) for human-annotated SFT. 
In light and dark blue are statistically significant (r. $p$<0.05 and $p$<0.001) results under a McNemar's test, when compared to ICL. 
This approach provided reasonable-to-good improvements across all locales except Geordie. 
These results must be interpreted with caution due to sample size. 
A detailed breakdown is in the Appendix, in \tabref{allmcnemar_ft}.}
\label{tab:ft_vs_baseline}
\end{table*}

\section{Ablation Studies}\label{sec:ablation}

We briefly discuss our studies on linguistic characteristics (\secref{ablationtypology}), a per-task analysis (\secref{ablationpreds}), and the impact of rubric decomposition (\secref{ablationrubric}). 
See \appref{detailedresults} for details.

\subsection{Impact of Linguistic Characteristics}\label{sec:ablationtypology}

\subsubsection{Relationship to Linguistic Distance}
\begin{figure}[]
\centering
\includegraphics[width=0.99\linewidth]{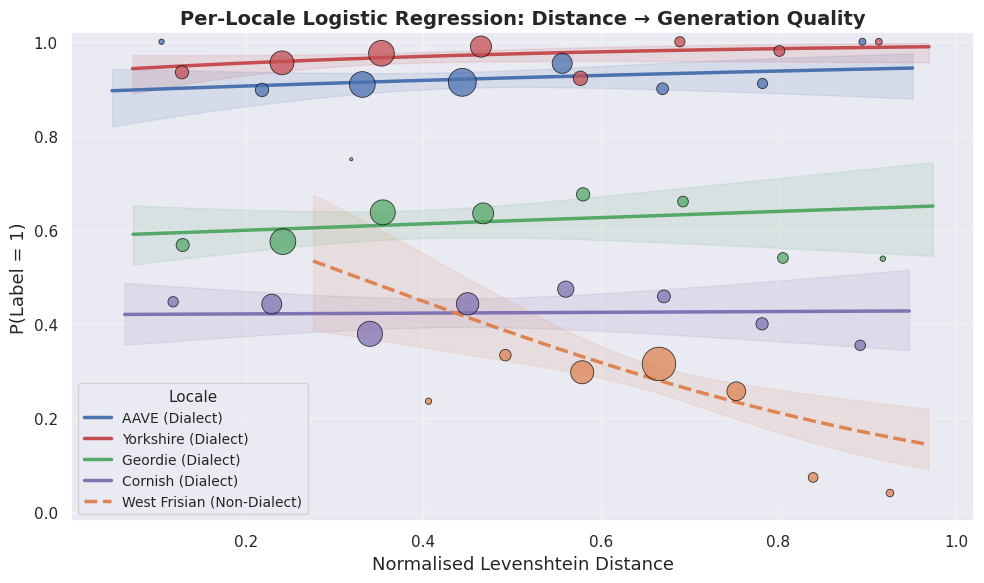}
\caption{Levenshtein distance versus LLM performance per locale. 
A logistic model shows that distance negatively predicts the odds of label=1 in West Frisian, but not in English dialects ($p$ < 0.001).}
\label{fig:distances}
\end{figure}

Computing distance per locale is challenging since, to our knowledge, no English-specific distance methods exist for dialects which also fit West Frisian. 
Hence we followed standard practice \cite{10.1093/oxfordhb/9780199922765.013.0058} and measured it with Levenshtein distance. 
Then we correlated the distance between the human-transcreated locale prompt and the original en-US-std prompt with their label=1 percentage per LLM. %
A logistic model showed that Levenshtein distance is a negative predictor of generation quality in West Frisian ($\beta = -2.91$, $p < 0.001$): as this distance increases, the odds of producing a label=1 output drop by roughly 94\% (\figref{distances}). We did not find significant linear correlations for the dialects.

\subsubsection{Relationship to Resourceness} 

Due to the complexity on measuring the volume of available online data for our locales, we estimated speaker numbers and used these as proxies for resourceness. 
Population sizes are still difficult to estimate, although from what we observed there was a relationship between resourceness and the pattern in dialects. 
In descending order of baseline LLM performance (from \tabref{baselineresults}) and estimated resourceness:

\noindent\textbf{AAVE:} more than 30 million urban and `several million' rural speakers \cite{ewave-15,ewave-16}. \\
\textbf{Yorkshire:} less than 5.48 million. This is taken from the 2021 census population of the Yorkshire \& Humber region \cite{ukcensus}. It is widely used as the dialect’s absolute ceiling, as the census does not track dialect use. \\
\textbf{West Frisian:} 470,000 speakers in the main province of the Netherlands \cite{lewis2009ethnologue}. \\
\textbf{Geordie:} approximately 800,000 speakers \cite{Watt_Allen_2003}. \\
\textbf{Cornish:} between 99,754 and 570,300 speakers. The lower bound comes from people who self-identified as having Cornish national identity in the 2021 census \cite{ukcensus}. The upper bound is the number of residents of Cornwall in 2021. It is unlikely that a significant portion of people outside of the region is an everyday speaker of Cornish English \cite{ukcensusid}.

\subsection{Per-Task Prediction Analysis}\label{sec:ablationpreds}

We performed a brief analysis before and after fine-tuning on LLM performance per task. 
On average, OpenCode was more challenging for the LLMs than other tasks, with an average -7.2 F$_1$ drop with respect to the mean, mainly driven by drops in Cornish and West Frisian (-11.0 for both). AC1 likewise dropped by -0.08. 
The best-performing split was GSM8K, with +4.0 and +0.05 increases in F$_1$ and AC1, respectively, and an average AC1 of 0.59. %
After fine-tuning, Qwen3's performance increased noticeably in OpenCode (+14.6\%, +3.8, +0.3 accuracy, F$_1$, and AC1), but SHP had drops of -4.9\%, -3.1, and -0.08 in all locales. 
That said, most changes were not significant: AAVE and Yorkshire had only OpenCode at $p$< 0.001; West Frisian had OpenCode and Wildchat at $p$ < 0.05; and Geordie had none. 
Improvements in Cornish were $p$ < 0.05 save for OpenOrca for strategy 1 and GSM8K and OpenCode for strategy 4. 

\subsection{Impact of Rubric Decomposition}\label{sec:ablationrubric}

We probed (1) the per-criterion accuracy of LLMs (strategies 4 and 5); and (2) the impact of the choice of aggregation function from \secref{annotation}. %
Strategies 4 and 5 followed the same pattern as 1-3: providing reasons was slightly lower-performing than not (-2.2\%, -2.6 accuracy and F$_1$); and shots provided minimal improvements, with positive OLS slopes of 1.9 and 1.2. Accuracy, F$_1$, and AC1 aligned with the human-human agreement on quality. 
When altering the criterion for label=0 from requiring at least one zero to three or more, Yorkshire and AAVE's scores increased, but decreased in Cornish and West Frisian. 
Ensembling the models per criterion through a majority vote negatively impacted performance, and requiring full agreement was most effective. 
Note that these approaches are incomparable with the rest of our results.

\begin{figure*}[]
\centering
\includegraphics[width=0.195\linewidth]{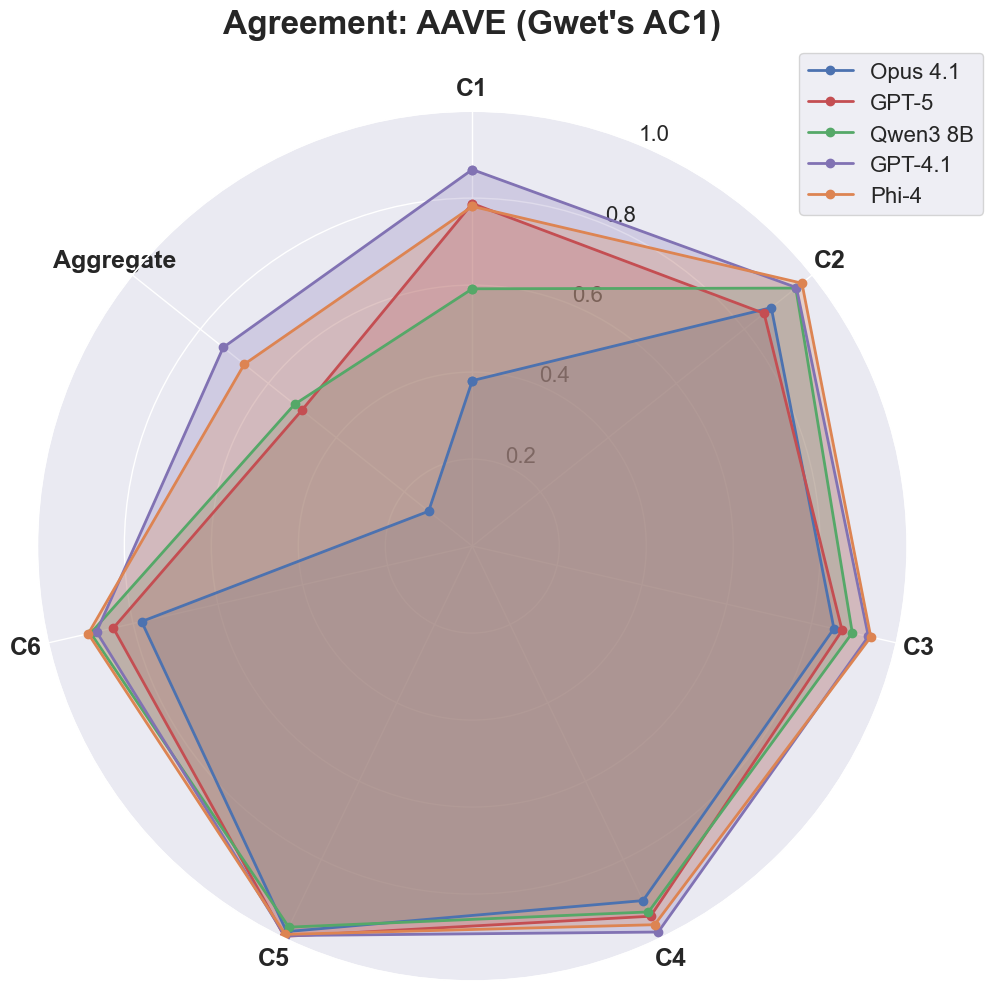}
\includegraphics[width=0.195\linewidth]{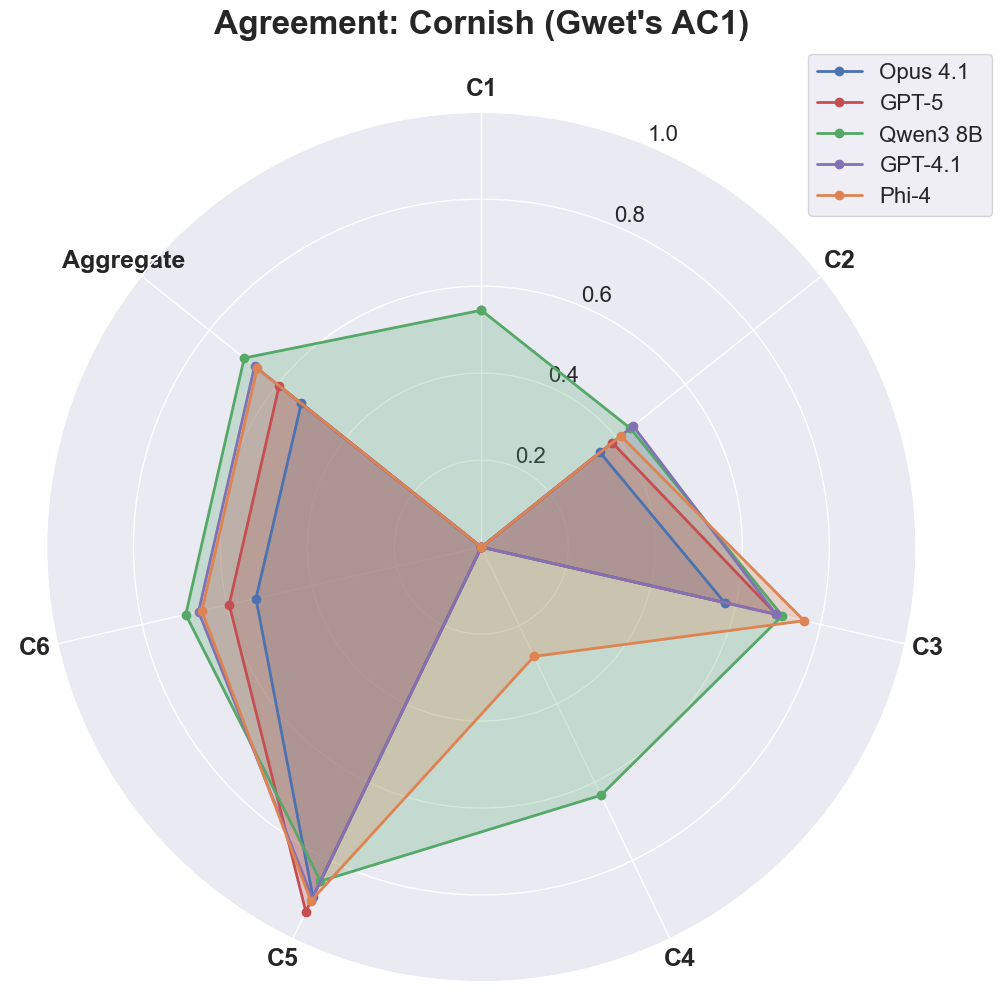}
\includegraphics[width=0.195\linewidth]{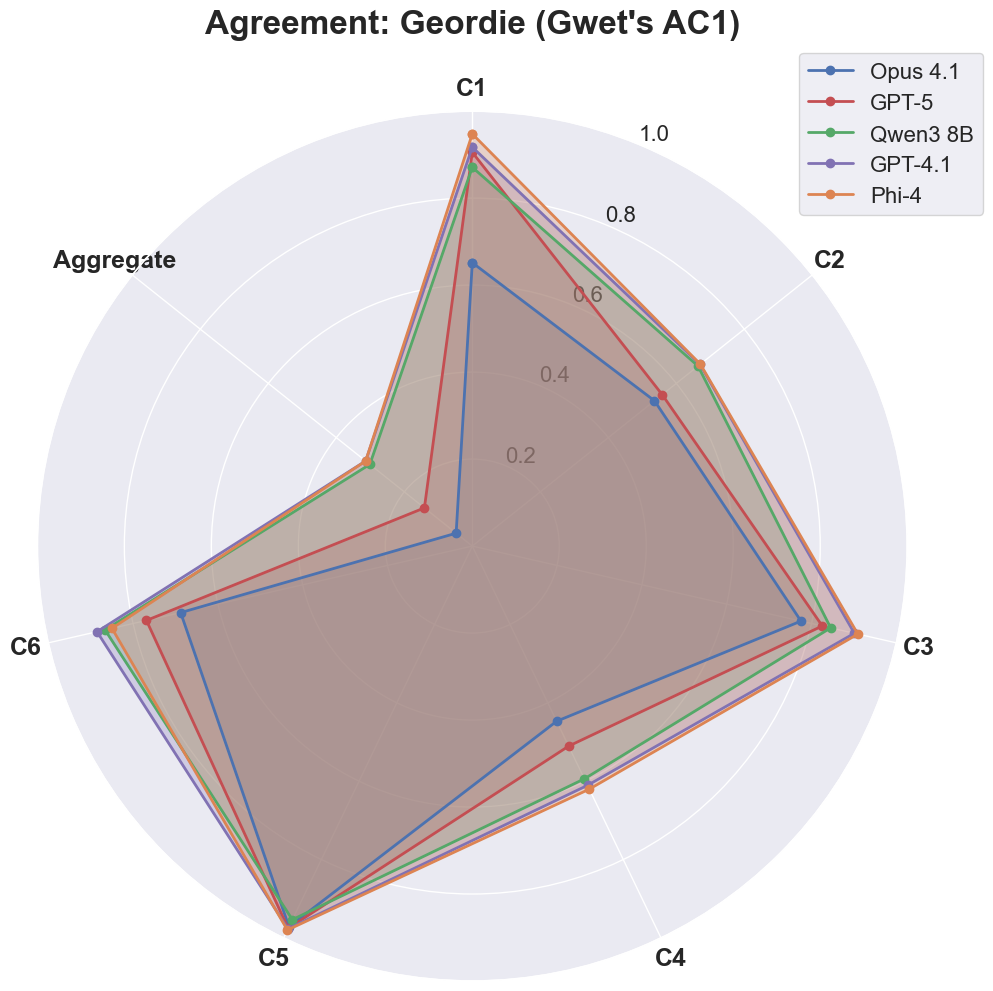}
\includegraphics[width=0.195\linewidth]{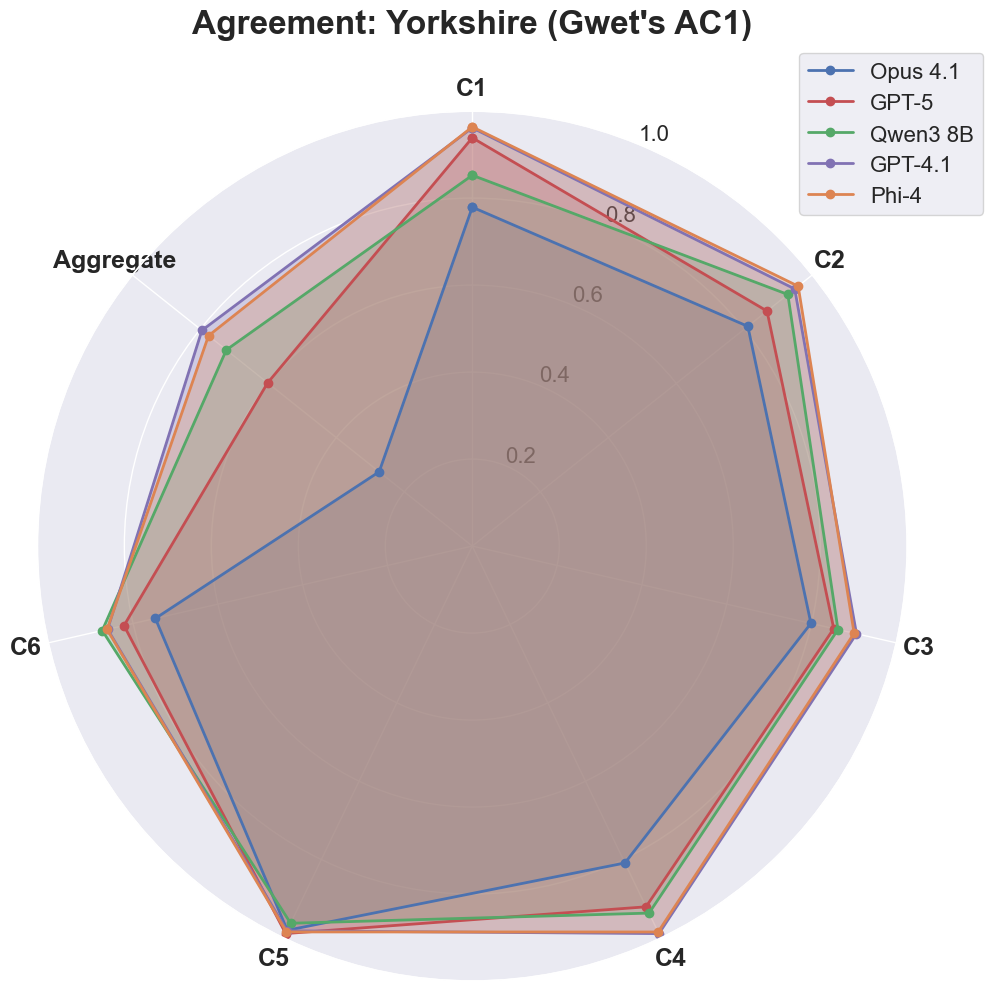}
\includegraphics[width=0.195\linewidth]{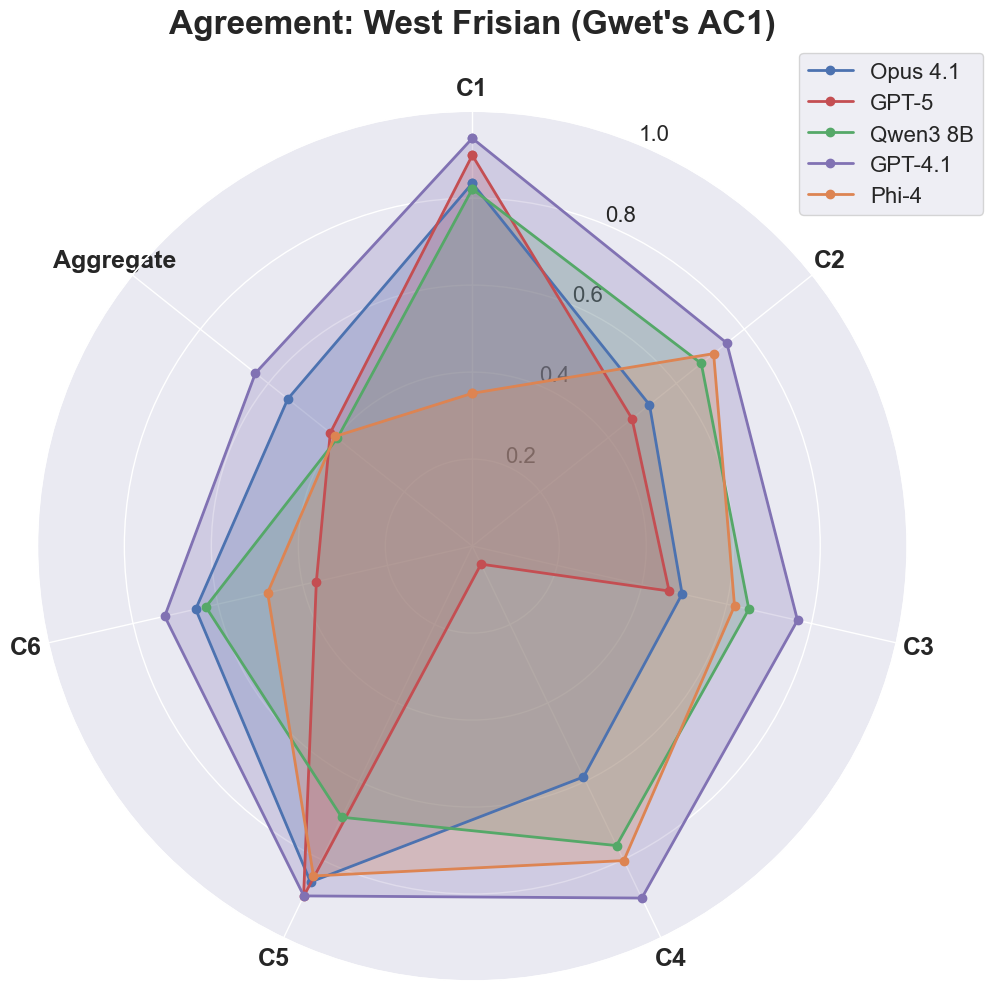}\\ 
\bigskip
\includegraphics[width=0.195\linewidth]{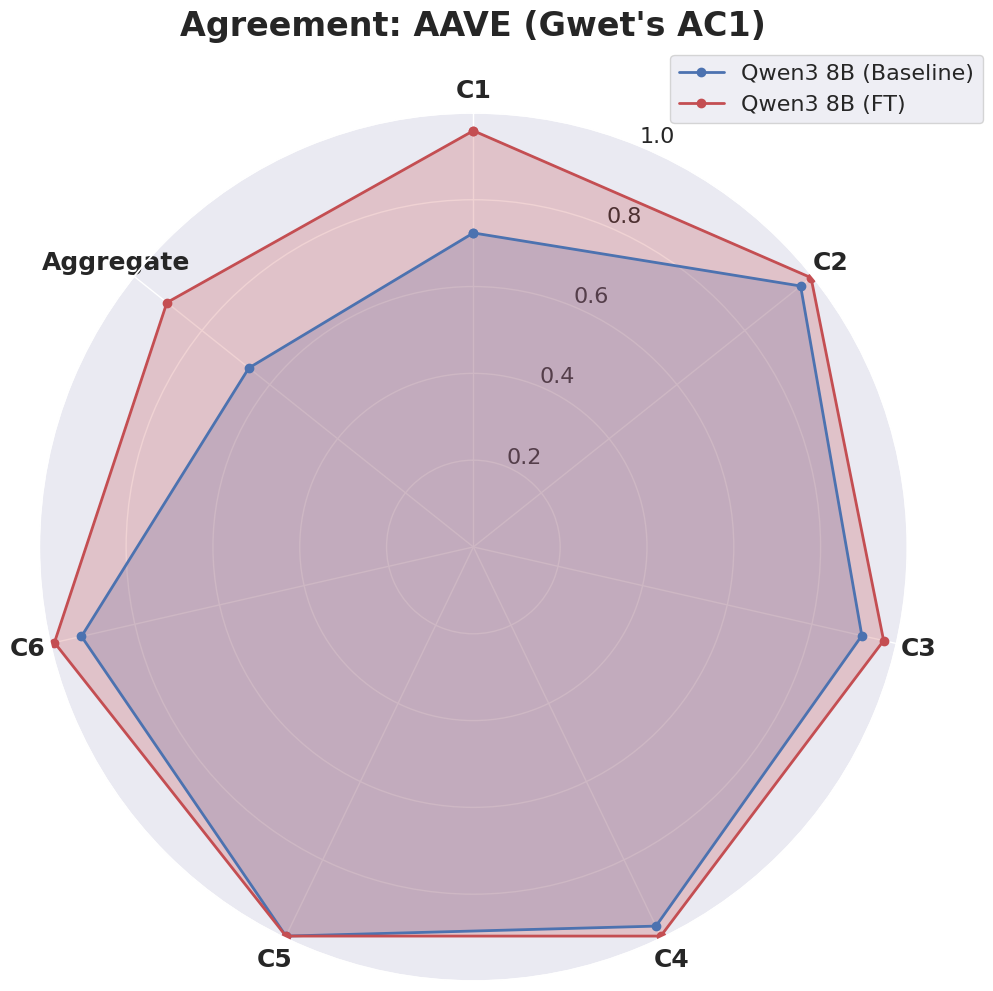}
\includegraphics[width=0.195\linewidth]{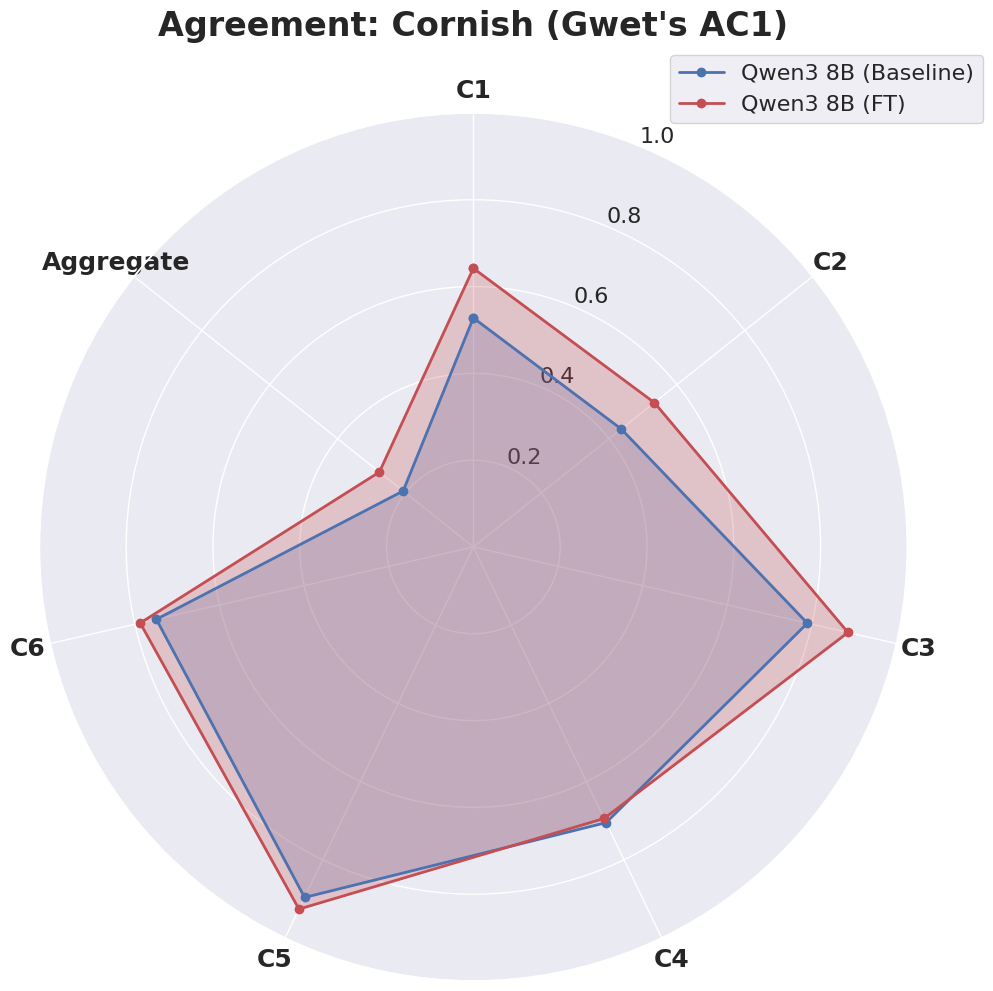}
\includegraphics[width=0.195\linewidth]{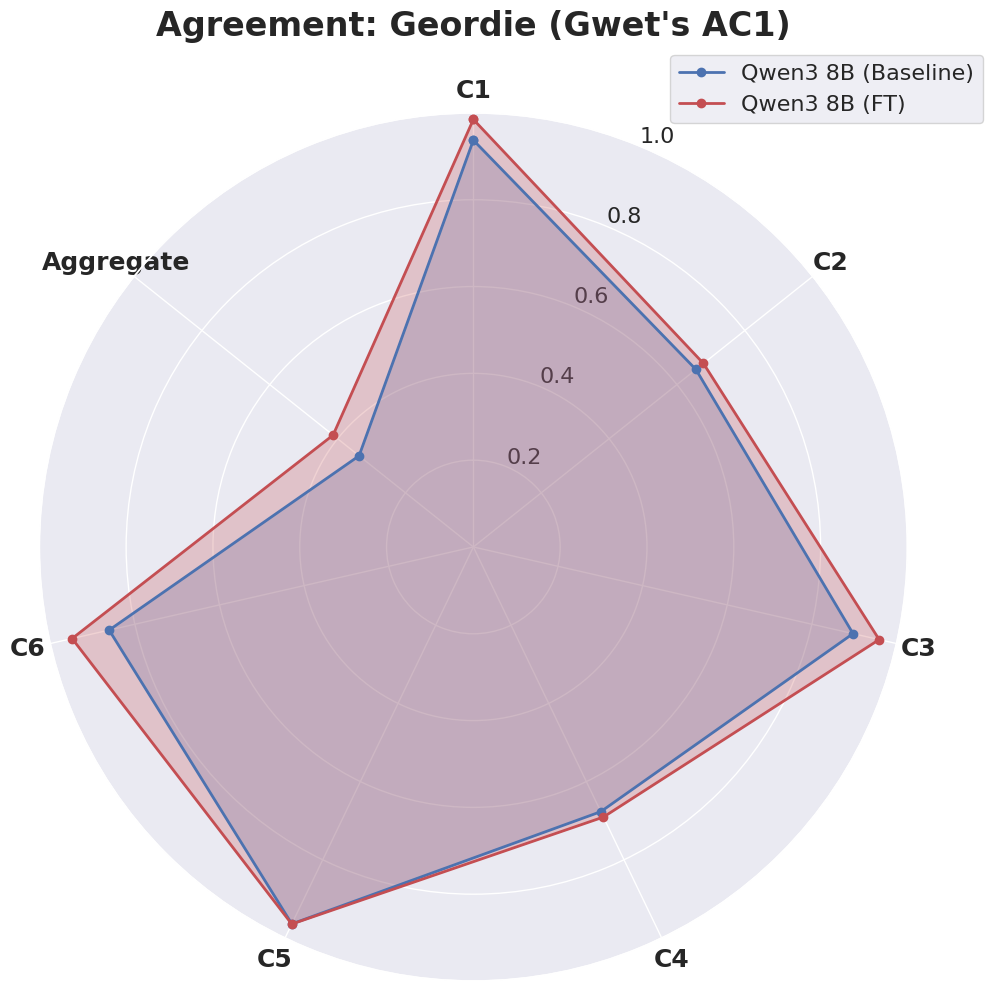}
\includegraphics[width=0.195\linewidth]{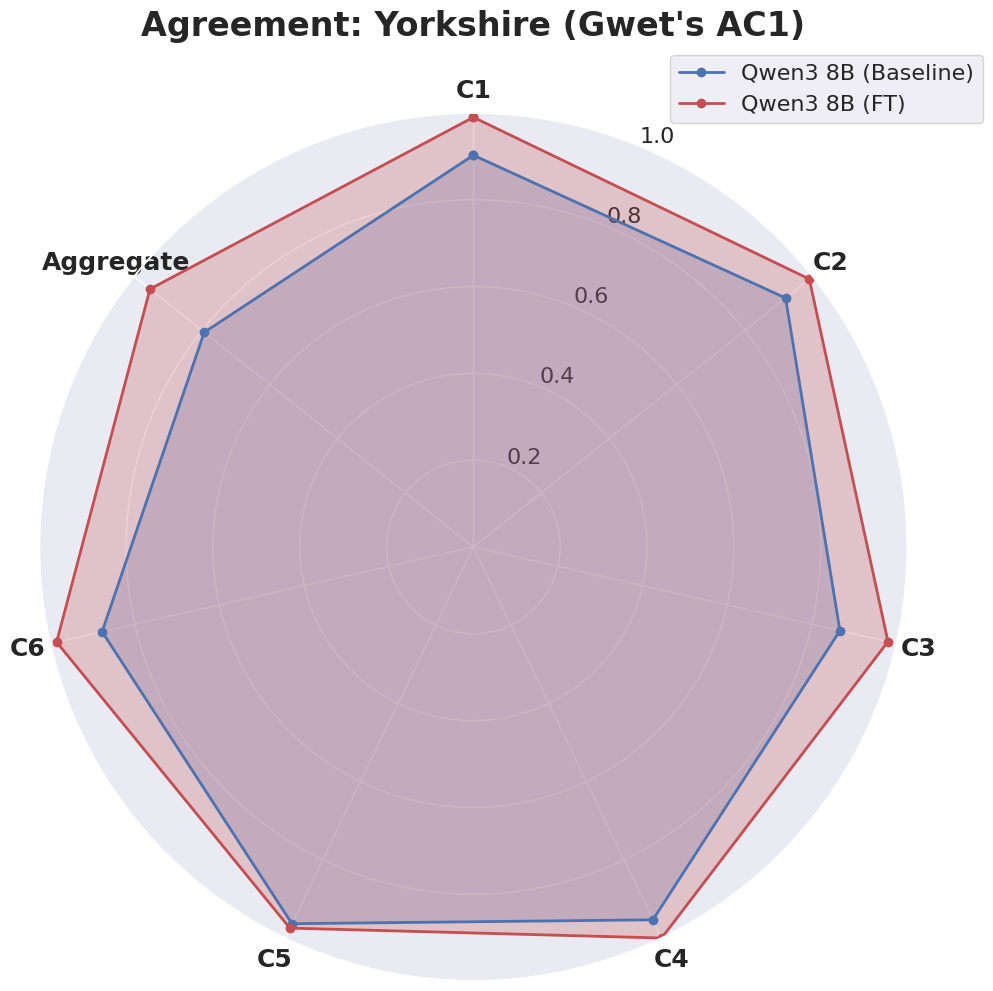}
\includegraphics[width=0.195\linewidth]{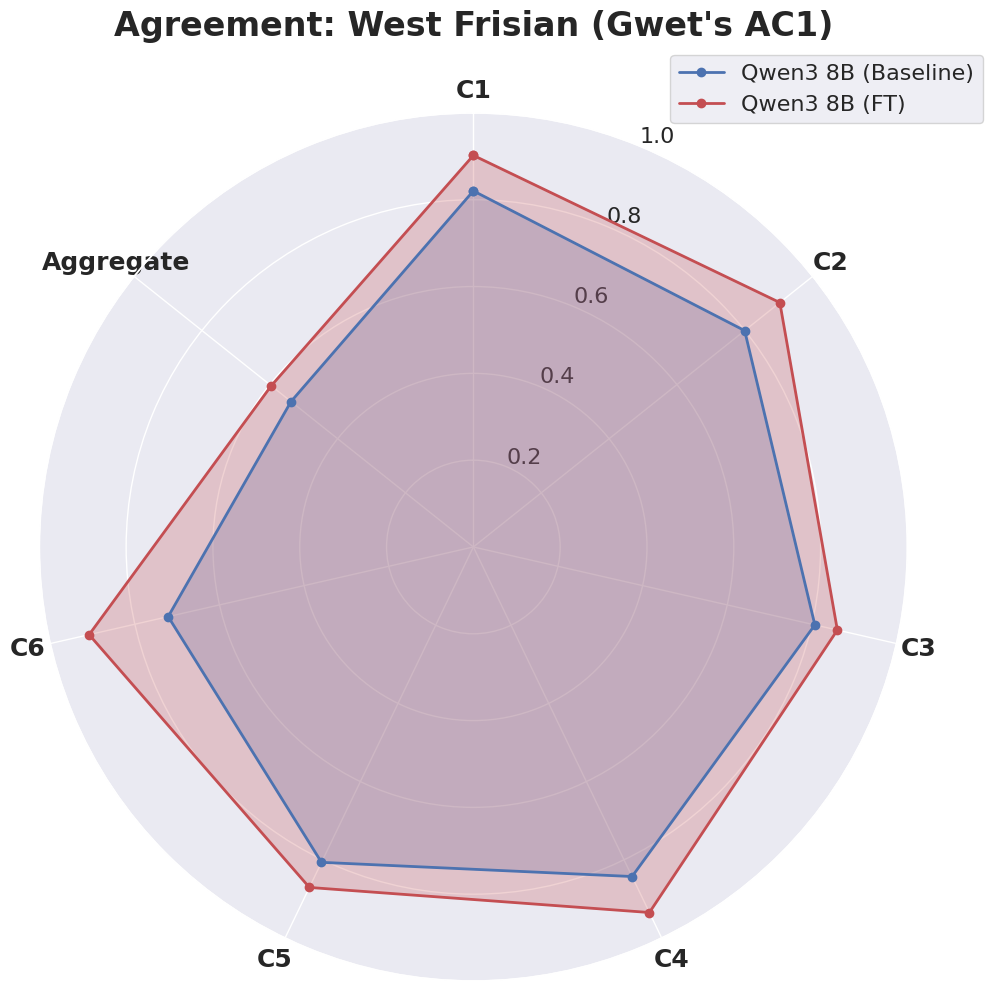}
\caption{Strategy 4 LLM-human agreements for \textit{(left-to-right)} AAVE, Cornish, Geordie, Yorkshire, and West Frisian. 
\textit{Top}: agreements during ICL (all LLMs). 
\textit{Bottom}: agreements for fine-tuned Qwen3. 
During ICL, LLM-human AC1 approximated the human-human AC1 distribution: high AC1s corresponded to locales with high human-human AC1 (AAVE, Yorkshire), and symmetrically for low AC1 (Cornish, primarily). LLM performance as measured by F$_1$ also followed this pattern. 
After fine-tuning, AC1 increased, although in dialects these improvements largely followed the pattern. 
West Frisian had low AC1, but presented noticeable improvements.
}
\label{fig:baselineagreement}
\end{figure*}

\section{Discussion}\label{sec:discussion}

\subsection{Generation of TSD}

LLMs were consistent in their performance distribution across locales. All LLMs generated more content with label=1 in Yorkshire and AAVE than in Geordie, Cornish, and West Frisian. 
This was regardless of magnitude: Qwen3's outputs were the lowest-rated in AAVE, yet these were still higher quality than its outputs in Geordie, Cornish, and West Frisian. 
The fact that AAVE and Yorkshire were the highest-rated locales was surprising: Geordie is much closer linguistically to Yorkshire, and yet the average quality gap between these two was 47.1\%. 
Our error analysis suggests that LLMs may only be used to generate task-specific data in certain dialects, but not all. Likewise, they must be calibrated ahead of time, given the disparity between closely-related locales. 
Remark that the high quality observed in AAVE seemingly contradicts earlier works which have found that LLM responses for this dialect. We note, however, that our generation prompt was especially designed to respond in the locale, something not common in live scenarios. 
Indeed, Opus-4.1's low performance in AAVE aligns with known results \cite{beaglehole}. 
Nonetheless, we can conclusively indicate that \textbf{this approach is an effective data-expansion method for AAVE and Yorkshire English}. %

\subsection{Evaluation of TSD}

Human-LLM AC1 and F$_1$ scores in ICL were largely aligned with human-human AC1: that is, the performance of LLMs during judgement followed the same distribution as the human-human agreement. 
Fine-tuning kept the pattern in dialects, too, but not in West Frisian: strategies 1 and 4 improved in almost all locales, yet the gap between high- and low-performing dialects remained. 
Remark that human-LLM agreement is--in theory--supposed to be near 1.0, even when human-human agreement is not. 
This is because they measure different things: good agreement between humans and LLMs implies that the latter concur with the human \textit{consensus}. %
Human-human agreement measures to which extent the consensus holds. 

The above raises two questions: (1) why are dialects, but not West Frisian, subject to that pattern? and (2) why does the pattern occur to begin with?

For the first question, %
recall our finding that \textbf{linguistic distance is an ineffective predictor of generative quality in dialects}, but not for West Frisian. 
Highly divergent distributions (e.g., languages) have distinctive patterns (grammars) which LLMs can handle better, whereas subtle dialectal variations introduce more ambiguity. 
This is also supported by the fact that there is no indication that Qwen3 was trained in West Frisian; and by its low label=1 proportion and AC1 in ICL, but large improvements when compared to other dialects after fine-tuning (\tabref{allmcnemar_ft}). 

Our linguistic distance hypothesis also partially answers the second question. Our resourceness estimates provide the other part, since they correlated with the observed human-human agreement in dialects, but not in West Frisian. 
The latter, however, has official status in the Netherlands and likely has more online presence. 
Hence, it follows that \textbf{an LLM's ability to judge content in a specific locale is based on its exposure to it}. %

Explaining the pattern via resourceness goes beyond the seemingly obvious statistical fact that more data implies better performance: many English dialects are low-resource. With few annotators available, there will be less agreement on what a standard dialect--including spelling--comprises, and thus \textbf{data will have higher variances in human quality judgements}. This translates into low agreement, and hence into low performance due to \textbf{the LLM's inability to learn to identify deep-yet-subtle linguistic cues as required by our rubric}. 

Our fine-tuning efforts support this hypothesis. 
Recall DPO's failure in Cornish: this outcome is reasonable, as DPO is known to not be effective at injecting new knowledge, unlike SFT \cite{thakkar-etal-2024-deep}. 
However, SFT's main drawback is the need for data, and synthetic data expansion was only effective for dialects where the LLMs were \textit{already good at it} during ICL. 
It follows that synthetic data approaches for dialects strongly depend on the quality of its generators; with \textbf{low performers doing more harm than good}, as observed in Yorkshire and Cornish. 
Note as well that, although strategy 4 was not as effective as strategy 1 during ICL, after fine-tuning it matched it, as well as the pattern. 
Given that these are, for all intents and purposes, different identification problems, it follows that, in general, \textbf{LLM judgement of linguistic cues is affected by this pattern}.

In sum, we argue that the pattern stems from resourceness and the ambiguous lexical nature of dialects. 
Its existence, in turn, impacts the feasibility of adapting LLMs for low-resource dialects: \textbf{they become ineffective judges since they are unable to gain the ability to identify deep linguistic structures} (e.g., quality, grammaticality) with limited data and no prior exposure.

\subsection{Alternate Explanations and Limitations}
As noted, resourceness could impact variance in speaker judgements, and thus human-human agreement. Adding more annotators could show results different to the pattern observed. 
Likewise, the small sample size from our evaluations, although realistic within our scarce-data setup, could mean that some statistical conclusions are not as robust. 
However, our main assumption is that low-resource dialects lack readily-available human annotators. 
Hence, it is reasonable to assume that (a) practitioners will also lack sufficient annotators, and thus high-agreement data; and (b) they will need to rely on metrics beyond accuracy and F$_1$ to draw meaningful conclusions in small datasets. 
Our findings thus focus precisely on this: we indicate low agreement leads to low performance, and it is likely that LLMs are subject to this pattern and might underperform. %
Further limitations are in \secref{limitations}.

\section{Conclusion}\label{sec:conclusion}

We explored the relationship between dialect-specific data quality and its feasibility to improve LLMs for these locales. 
We found that the distribution of human-human agreement when judging the quality of TSD was replicated by other measurements, namely LLM-human agreement and LLM-as-a-judge scores. 
This is non-intuitive because they measure different things: LLM-human agreement measures an LLM's alignment with human consensus; while human-human agreement measures the frequency by which that consensus was adopted by all parties. 
Further experiments showed that this pattern was invariant to fine-tuning--if anything, some approaches widened the gap. 

Upon closer scrutiny we realised that West Frisian, our control, did not fully adhere to the pattern: low LLM-as-a-judge performance and generative quality was largely improved with fine-tuning. 
Hence, we argued that the pattern could be explained through two factors: (1) resourceness, where availability and human-human agreement could impact both data quality and LLM exposure; and (2) linguistic characteristics distinguishing English dialects from West Frisian. 
It then follows that LLMs-as-judges might be incapable of grasping deep linguistic structures under data-scarce and no-prior-knowledge scenarios. 

Even though edit distance is a coarse proxy for true linguistic divergence, it allowed us to explain the pattern. Hence, in these situations, we argue that it is not viable to fully rely on statistical methods to improve low-resource dialectal coverage. Given our findings on linguistic distance, we argue that it is needed to develop \textbf{effective linguistics-informed tools}. 
The existence of the pattern itself likewise emphasises the need to involve native speakers as much as possible in these pipelines. 

On the other hand, recall that we also found signals on the feasibility of scaling TSD for AAVE and Yorkshire; as well as West Frisian's excellent improvement after fine-tuning. Hence, we argue that the path to improving and building fair and inclusive LLMs is difficult, but not unattainable.

\newpage
\section*{Limitations}\label{sec:limitations}

Our work is limited in the number of locales covered, and on the longevity of the LLMs used. 
For the first limitation, we noted in \appref{litreview} that some of the locales from our work have not been studied in the context of LLMs--at least not within the timeframes reviewed. 
Still, much remains to be done to improve coverage, particularly for UK dialects and other severely endangered Frisian languages, such as Saterland. 
The second implies that some of our results--particularly these around LLM performance--could be outdated with newer releases. 
However, that is precisely the goal of our work: to provide evidence by how and by which means LLMs can improve on certain dialects. 
With that in mind, we have released the full suite of artifacts from this work. 

\section*{Ethical Considerations}\label{sec:ethics}

All aspects of our research were reviewed and approved by our institutional review board. 
Annotators were contracted via an external annotation company and compensated based on seniority, starting at a rate of 25.4 USD/hr. 
AI assistants were used to generate the code to create the figures in this work. The code was reviewed and modified as needed by the authors.

Our work focuses on improving fairness and linguistic preservation, and thus we do not believe that it could be misused. 
Nonetheless, there are possible unintended side-effects of our resources if not used carefully. 
In particular, it could be that practitioners developing systems use our corpora for other dialects without engaging first with native speakers. 
This could risk misrepresentation and propagation of stereotypes. 
We strongly believe that our our work has more value when open-sourced and made freely available to the community with the purposes of improving fairness and linguistic preservation. 
Hence, we will release the full suite of code and data used in this work under a MIT licence.
However, we also strongly encourage practitioners to employ caution when using these resources, and vet them prior to release.

\DeclareRobustCommand{\DE}[3]{#3}
\bibliography{biblio}

\DeclareRobustCommand{\DE}[3]{#2} %
\appendix

\section{Related Work}\label{app:background}

\subsection{Dialect-Based Evaluation of LLMs}

Even though it is well-known that multilingual performance declines with a language's online availability \cite{hasan2024large,hada2023large}, multilingual dialect and variety research is usually overlooked in favour of the standardised version of a language \cite{blasi-etal-2022-systematic,greenaave,
de-wynter-2025-awes,faisal-etal-2024-dialectbench}. 
The existing evaluations frequently show that task-specific performance of LLMs in these is markedly lower than in the standard version of the language (e.g., \citealt{pan-etal-2025-analyzing,zhou2025dialectgenbenchmarkingimprovingdialect,faisal-etal-2024-dialectbench,
faisal-anastasopoulos-2025-testing, 
groenwold-etal-2020-investigating,deas-mckeown-2025-artificial}, to name a few). 

An in-depth review of this area may be found in \appref{litreview}. 
Amongst our findings we note that, out of all the papers covering dialects and varieties between 2024 and the end of 2025, \textbf{no paper studies British English dialects, and only three explicitly studied West Frisian}, in the context of LLMs. 
Instead, the most-studied language is Arabic (54\% of the works), followed by English (19\%), primarily under AAVE and Singaporean. 
In addition, it is relatively rare for studies in this area to perform modelling work and often opt to release corpora instead: although most (65\%) papers claimed to release some artifacts, only 11\% released a model, and 13\% contained broken links.

\subsection{Dialect Adaptation Research}

The SOTA for dialect and variation adaptation typically involves fine-tuning and model merging \cite{khade2024challenges}, data augmentation (usually with synthetic datasets, \citealt{faisal2024data, alexandrov2024bggpt, shang2024atlas}), prompt-based methods \cite{cahyawijaya2024llms}, or a combination thereof. 
The lack of labelled data is a hindrance. 
LLMs-as-judges are commonly used to scale in extremely low-resource scenarios, but multiple studies have documented reliability issues with LLM-based evaluations, including low agreement with human evaluations \cite{hada2023large, schroeder2024can, thakur2024judging}, and trustworthiness and cultural appropriateness of the source data \cite{taguchi-etal-2025-languages,sitaram-etal-2025-multilingual} and output \cite{dorn2025reinforcingstereotypesangeremotion}, thus demanding rigorous human validation. 

The most prominent work in dialect-based evaluation is arguably DialectBench \cite{faisal-etal-2024-dialectbench}. This comprehensive dataset, covering 281 dialects and varieties, is meant to benchmark model performance in traditional NLP tasks. 
The original paper and a follow up by \citet{faisal-anastasopoulos-2025-testing} both showed LLMs underperformed across the board. 
However, DialectBench does not cover the English dialects we cover, nor it is user-task oriented.

\section{Comprehensive Background}\label{app:comprehensivedialects}

\subsection{On Dialects}

There is no broadly accepted definition of a dialect with respect to a language, with its distinction typically muddled by geopolitics or cultural considerations. Even the denomination `dialect' could be considered offensive to some speakers \cite{lewis2009ethnologue}. 
This likewise applies to the convention as to what a `standard' language is. 
In the case of English, there has not been a (prescriptive) standardisation, and due to its pluricentric nature, multiple locations consider a given version their own, with specific othography, lexicon, and phonology: consider, for example, the word `pavement' (UK Standard, in England and Wales), `sidewalk' (General or Standard American), and `footpath' (General or Standard Australian). 
Making clear-cut distinctions within a continuum, such as within the Romance continuum (e.g., Spanish versus Portuguese) or, in the converse, within the Arabic macrolanguage (e.g., Moroccan and Gulf Arabic), is also challenging and influenced by the same factors \cite{crystal}. 
In this work, for classification purposes, we follow the conventions in Ethnologue,\footnote{\url{https://www.ethnologue.com/}} but overridden by native speaker preference when applicable (\appref{litreview}). 
For notation purposes consider all locales studied (save for West Frisian) dialects, as opposed to varieties, to further emphasise its cultural nature.

\subsection{Dialects Covered}

In this section we cover in detail the English dialects from our work. 
Our focus is in grammatical and lexical features. It must be noted, however, that all of them have very distinctive phonological characteristics which impact their writing conventions. 

\paragraph{AAVE:} AAVE is spoken in United States and parts of Canada primarily by African Americans and some Black Canadians. 
It is a dialect (variety) of African American English, the collection of dialects spoken by Black people in North America, which also includes other dialects such as African-American Appalachian English or African Nova Scotian English \cite{mufwene2,greenaave,10.1093/oxfordhb/9780199795390.001.0001}. 
Although specifics around its origin are less clear than in the other dialects \cite{developmentaave}, AAVE is known to share grammatical structures present in other Southern American English dialects, such as Southern White Vernacular English \cite{cukoravila,mufwene}. 
It is often used in daily, casual speech \cite{mufwene}. 
Morphosyntactically, characteristics unique to AAVE are the use of `be' for denoting invariants, aspect, and its possible omission in the present tense (`they walking too fast': they are walking too fast; `I be looking for you': I'm always looking for you), the use of `done' as a perfect tense (`you done changed': you have changed\footnote{\citet{greenaave} distinguishes `d\textschwa n' and `done' as distinct forms, such as in `I d\textschwa n done all my homework'.}), and relatively looser word order, especially in negations (`don't nobody go'; \citealt{greenaave}). 
Many terms originally from AAVE have been adopted in standard US English (e.g., `cool', `finna', or `cap').\footnote{\url{https://en.wiktionary.org/wiki/Category:African-American_Vernacular_English}, accessed 22 December 2025}

\paragraph{Yorkshire:} Yorkshire English (alternatively, Yorkie or Tyke) is a collection of English English dialects (North, Craven, Sheffield, and West; \citealt{lewis2009ethnologue}) primarily spoken in the Yorkshire area of the UK. 
Due to York being in the the Danelaw (specifically, the Kingdom of Jórvík; what is now Yorkshire), its dialect has noticeable Old Norse and Middle English (Northumbrian) influences \cite{oldnorse,yorkshireviking}. 
In terms of grammar and morphology, it has various characteristic to Northern English\footnote{There is no widely accepted definition on what some geographical regions of England, such as North and West Country, are.} and to the dialect itself, such as the Northern Subject Rule (e.g., adding an -s after certain pronouns in present tense) and multiple negatives. It is also known for very distinctive grammar, such as extra pronouns (`tha': you; `-yen': -one) and phrases (`I'll stand drop o'York': it is unbelievable; \citealt{yorkshireviking}).

\paragraph{Geordie:} Geordie is spoken primarily in the northeast of England, in and near Tyneside. 
It retains many archaic features characteristic of the Northumbrian dialect of Middle English and Scots, and--it being close to the border of the Danelaw--Old Norse \cite{oldnorse}. 
Some lexical features are shared with Yorkshire and other Northern and Scottish dialects, and others are unique to Geordie, such as `clamming' (hungry), `wor lass' (girlfriend), and `gan hyem' (go home) \cite{northumbriageordie,englandsnortheast}. 

\paragraph{Cornish (West Country):} Unlike Yorkshire and Geordie, the West Country dialect is a continuum spoken in the eponymous region of South West England (e.g., Cornwall, Devon, Somerset, Bristol). 
Like its neighbouring varieties, Cornish (the English dialect) is more influenced by Late West Saxon (the main language Beowulf is written in) and Cornish (the indigenous, now revived, Brittonic language). 
Given Cornwall's geographical location, Middle English was adopted later there than in the rest of England, making the dialect relatively new and distinct with respect to other West Country and English dialects.\footnote{See \cite{spriggs} for a list of reasons.} 

Morphosyntactically, in Cornish it is common to use `be' and to omit pronouns. 
Other features are double negatives, as well as archaic pronouns (`thee', e.g.), albeit the latter are becoming more common. 
Quintaessential Cornish examples such as `enow' (too much, plenty), `emmet' (ants, also a word for tourist), and `pop-an-towse' (a fuss, e.g. `t'was some pop-an-towse up square just now') \cite{cornishdict,cornishdict2}. 

\subsection{West Frisian}
West Frisian is a language used by approximately 470,000 people (as of 2001) in the northwestern part of the Netherlands, primarily in the province of Friesland (Frysl\^an) \footnote{\url{https://www.omniglot.com/writing/westfrisian.htm}, accessed 23 December 2025.}. 
It is considered a language with `exceptionally limited [online] resources', as per the taxonomy of \citet{joshi-etal-2020-state}. 
It is part of the Frisian language family, which also includes Northern Frisian and Saterland Frisian. 
The latter two are considered endangered by the \citet{atlasoflanguages}, with Northern Frisian having only approximately 10,000 speakers in 1970; and Saterland 2000 \cite{lewis2009ethnologue}. 
While itself is comprised of four major varieties,\footnote{\url{https://glottolog.org/resource/languoid/id/mode1264}, accessed 22 December 2025.} in this work we consider the standard version (Western Frisian). We do not make distinctions between its three major dialects (Klaaifrysk, S\'udhoeksk, and W\^aldfrysk): their differences are mostly phonological, with some minor lexical distinctions. 

Depending on the source, the Frisian languages--and West Frisian in particular--are considered the closest living relatives to English. The closest language, Scots, is a sister language also belonging to the Anglic family \cite{atlasoflanguages}. 
Due to its lack of official standardisation, however, we opt to work with West Frisian as our control. 
It is also worth noting that there exists a debate as to whether Old Frisian and Old English comprise a united phylogenetic family \cite{buczek,stiles}, albeit the consensus is that both languages share many features.

\section{A Survey on Works Covering Locales}\label{app:litreview}
As part of our work, we performed a brief literature survey of the major works studying dialects in (any) language, as well as West Frisian. 
For this, we queried the SemanticScholar API \cite{Kinney2023TheSS} for works between 1 January, 2024, and 18 December, 2025. 
For dialects, we used the query `dialect llm -dialectical -speech -asr' between 1 January, 2024, and 18 December, 2025. 
For West Frisian, we used `frisian', between 1 January, 2020, and 18 December, 2025. The expanded data window was due to the the low number of works found in the dialect-specific benchmark. 

After querying, we filtered the results to have as a primary field computer science, and manually reviewed based on the filtering criterion below. 

\begin{itemize}
    \item It must relate to \textit{text-based evaluation} (speech recognition, e.g., is excluded) of decoder-only LLMs. Surveys and reports from shared tasks are excluded. 
    \item It must clearly state which locales are evaluated, and provide a breakdown by locale of the results.
    \item If it is a paper released in 2024, it must have been peer-reviewed.
    \item In the case of West Frisian, we ensured that `Frisian' meant `West Frisian' and not the related languages Saterland and North Frisian, or the Dutch dialect spoken in Frysl\^an. 
\end{itemize}

The final number of papers were 74 for dialects, and four for West Frisian \cite{sitaram-etal-2025-multilingual,dewynter2025labellingdataunknownreferences,faisal-etal-2024-dialectbench,faisal-anastasopoulos-2025-testing}. 
In the latter, three papers released data and neither released artifacts. 
The resulting papers were once again manually reviewed to determine the artifacts released, with the criterion that these must be clearly stated within the works. We also verified the links provided. 
The results are in \tabref{datasurvey}. 
Outside of seven works, all papers focused on the dialects of a single standard language. 
Most works surveyed included en-US-AAVE (10 out of 14), and none included UK dialects. 

Out of the 74 papers, in terms of language distribution, there were 22 languages. 
Arabic was the most studied language (54\%), followed by English (19\%) and BCMS (6\%). 
In the 30 papers covering dialects of English, the most studied dialect was AAVE (30\%; nine), followed by Singaporean (13\%), and Nigerian and Indian (13\% and 10\%). 
All other dialects (Southern American, Chicano, Australian, Appalachian, Hong Kong, Irish, Canadian, Sri Lanka, and West African Pidgin) were only studied once. 
British dialects of English were not present in the works. 

The least covered languages were Vietnamese, Thai, Slovak, Dutch, Hindi, Spanish, Kurdish, Finnish, and Latvian, all at 1\% (that is, one single work) coverage. 
In terms of artifacts, 43\% released data, 33\% code, 11\% a model, and 35\% did not release either; with 13\% of the works containing broken links.

\begin{table*}[]
    \centering
    \small
    \begin{tabular}{l|p{0.5\linewidth}|p{0.1\linewidth}}
    \toprule
    \textbf{Work} & \textbf{Dialects} & \vtop{\hbox{\strut \textbf{Artifacts}}\hbox{\strut \textbf{released}}} \\
    \midrule
\citet{mousi-etal-2025-aradice} & \textbf{ar}: afb, arz, apc & data\\
\citet{hannani-etal-2024-assessing} &\textbf{ar}: ary & - \\
\citet{10651099} &\textbf{ar}: arz, afb, apc, ary, arq, aeb & - \\
\citet{moudjari-benamara-2025-dialects} &\textbf{ar}:  arq, aeb, apc, afb & models$^\dagger$\\
\citet{bouomar-abbas-2025-arasim} &\textbf{ar}: arz& code, data \\
\citet{demidova-etal-2024-arabic} &\textbf{ar}:arz, ae, jo, pl& - \\
\citet{10850941} &\textbf{ar}: arq, arz, afb, apc& - \\
\citet{10989347} &\textbf{ar}: arz, apc, afb, ary, arq, aeb, acm& - \\
\citet{robinson-etal-2025-al} &\textbf{ar}:arz, ary, arq, aeb, apc, afb, apd & code, data \\
\citet{alabdullah2025advancingdialectalarabicmodern} &\textbf{ar}: arz, apc, afb & code, data, models \\
\citet{almonef2025absherbenchmarkevaluatinglarge} &\textbf{ar}: sa (north, west, central, south, east) & data$^\dagger$ \\
\citet{shang-etal-2025-atlas} & \textbf{ar}: ary & model \\
\citet{bhatti2025mcqopenendedarabiccultural} & \textbf{ar}: arz, apc, afb, ary, arq, aeb & - \\
\citet{ibrahim-etal-2025-bridging} &\textbf{ar}: ary, arz & - \\
\citet{yakhni-chehab-2025-llms} &\textbf{ar}: lb & - \\
\citet{beidas} & \textbf{ar}: arz, apc, apd, aeb, arq, ary, afb, ayl, acx, ayh, ayn, acq, jye & - \\
\citet{ibrahim-2024-cufe} &\textbf{ar}: arz, ae, jo, pl& code$^\dagger$\\
\citet{altakrori2025dialectalarabicmmlubenchmarkingdialectalcapabilities} & \textbf{ar}: arz, ary, sa, sy, ae & data, models \\
\citet{aftiss-etal-2025-empirical} &\textbf{ar}: ary & code, data, models\\
\citet{hannani} &\textbf{ar}: ary & - \\
\citet{alharbi-etal-2025-evaluating} &\textbf{ar}: arz, ary, lb, sa & -\\
\citet{Qarah_Alsanoosy2025} &\textbf{ar}: ary & - \\
\citet{kanjirangat2025exploringdataparameterefficient} &\textbf{ar}: arz, afb, apc, ary, arq, aeb& -\\
\citet{ELBELTAGY2024296} & \textbf{ar}: arz & code, data\\
\citet{yakhni2025finetuningllmslowresourcedialect} & \textbf{ar}: lb & code, data \\
\citet{mahdi2025llmsunderstandtunisianarabic} &  \textbf{ar}: aeb & code, data\\
\citet{Alahmadi2025} &  \textbf{ar}: Saudi & - \\
\citet{magdy-etal-2025-jawaher} & \textbf{ar}: arq, arz, ayl, aeb, acm, qa & data $^\dagger$\\
\citet{alostad} & \textbf{ar}: kw & - \\
\citet{MAHOUACHI2025128260} & \textbf{ar}: aeb & data \\ 
\citet{shang-etal-2025-nile} & \textbf{ar}: arz & data, model\\
\citet{el-mekki-etal-2025-nilechat} & \textbf{ar}: arz, ary & model\\
\citet{10.3390/bdcc8110157} & \textbf{ar}: arz, ary, apd, sa, sy & - \\
\citet{alwajih-etal-2025-palm} &\textbf{ar}: afb, arq, abv, apc, ayl, acx, arz, apd, ary, aeb, abv, acm, km, dj, so, mr & code, data\\
\citet{barmandah2025saudidialectallamlorafinetuningdialectal} &\textbf{ar}: acw, ars & code \\
\citet{ayash2025saudiculture} &\textbf{ar}: sa & data \\
\citet{nacar2025uilevelevaluationallam34b} &\textbf{ar}: arz, apc, ary, acw, ars & - \\
\citet{10879471} & \textbf{ar}: sa & - \\
\citet{rtplx} &\textbf{ar}: ajp, arz, sa; \textbf{de}: gsw; \textbf{pt}: pt, br; \textbf{zh}: hans-cn, hant-tw; \textbf{ru}: uk; \textbf{bcms}: hr, sr & data \\ 
\citet{perak-etal-2024-incorporating} & \textbf{bcms}: hr-ckm, sr-tor; \textbf{sl}: cer & code, data \\
\citet{faisal2024data} & \textbf{bcms}: hr-ckm, sr-tor; \textbf{sl}: cer & code, data \\
\citet{faisal-anastasopoulos-2024-data} & \textbf{ckm}; \textbf{bcms}: hr-ckm, sr-tor; \textbf{sl}: cer & data\\
\citet{sami2025comparativeanalysisretrievalaugmentedgeneration} & \textbf{bn}: ctg, rkt, syl, Comilla, Habiganj, Tangail & - \\
\citet{prama2025llmslowresourcedialecttranslation} & \textbf{bn}: syl & data \\
\citet{10.1145/3701716.3715468} & \textbf{bn}: bd, West Bengal & - \\
\citet{10.1145/3715070.3749228} & \textbf{bn}: bd, West Bengal & - \\
\citet{bui-etal-2025-large} & 
\vtop{\hbox{\strut \textbf{de}: als, pfl, bar, stq, ksh, nds}
\hbox{\strut \textit{\textbf{Note}: the authors cover North Frisian as a `dialect' of German, but}}
\hbox{\strut \textit{it is a severely endangered language.}}} & code, data
\end{tabular}
    \caption{Papers evaluating LLMs on dialects between 1 January, 2024 and 18 December, 2025, sorted alphabetically by the nominal standard variety of the language. This table covers A-D. 
    When the work cites an explicit locale without corresponding ISO 639-3 code, we default to the locale. 
    We consider BCMS a supralanguage; and due to the doubtful classification of Čakavian, we follow the original authors and consider it a dialect of Croatian. 
    Note that the papers sometimes cover more languages than what is displayed here, namely, the standard varieties of these. 
    Artifacts marked with $^\dagger$ had, at the time of writing this work, broken links, empty repositories, or indications of open sourcing without links provided.}
    \label{tab:datasurvey1}
\end{table*}

\begin{table*}[]
    \centering
    \small
    \begin{tabular}{l|p{0.5\linewidth}|p{0.1\linewidth}}
    \toprule
    \textbf{Work} & \textbf{Dialects} & \vtop{\hbox{\strut \textbf{Artifacts}}\hbox{\strut \textbf{released}}} \\
    \midrule
\citet{10.1145/3706468.3706496} & \textbf{en}: us-aave & - \\
\citet{otmakhova2025flukelinguisticallydriventaskagnosticframework} &\textbf{en}: sg& code \\
\citet{faisal-etal-2025-dialectal} & \textbf{en}: us-southern, us-chicano, us-aave, us-app, au, ng, sg, cn-hk, in, ie; \textbf{ar}: aeb, apc, ars, ary, arz, \textbf{az}: north, south; \textbf{lv}: east; \textbf{no-nn}; \textbf{ku}: northern; \textbf{bn}: dhaka; \textbf{fi}: pohjois-satakunta, keski-karjala, kainuu, etela-pohjanmaa, etela-satakunta, pohjois-savo, pohjois-karjala, keski-pohjanmaa, kaakkois-hame, pohjoinenkeski-suomi, pohjois-pohjanmaa, pohjoinenvarsinais-suomi, etela-karjala, lansi-uusimaa, inkerinsuomalaismurteet, lantinenkeski-suomi, lansi-satakunta, etela-savo, lansipohja, pohjois-hame, etelainenkeski-suomi, etela-hame, perapohjola & code \\
\citet{deas-mckeown-2025-artificial} &\textbf{en}: en-aave & code$^\dagger$ \\ 
\citet{ xu-etal-2024-beyond-perplexity} &\textbf{en}: en-aave & code \\
\citet{abbas2025smallscaledatapoisoningexacerbate} &\textbf{en}: en-aave, us-southern & code$^\dagger$\\
\citet{atwell-etal-2024-combining} &\textbf{en}: us-aave & code $^\dagger$ \\
\citet{zhou2025disparitiesllmreasoningaccuracy} &\textbf{en}: us-aave & code, data\\
\citet{srirag-etal-2025-predicting} &\textbf{en}: in, ng & - \\
\citet{mire-etal-2025-rejected} & \textbf{en}: us-aave & code, data\\
\citet{held2025relativescalinglawsllms} &\textbf{en}: ca, sg, lk, ng & code, data, models \\
\citet{coggins-etal-2025-aint} &\textbf{en}: us-aave, wape & data$^\dagger$\\
\citet{ryan-etal-2024-unintended} &\textbf{en}: in, ng & code, data\\
\citet{ng2024talkingyoutranslatingcodemixed} &\textbf{en}: sg & data$^\dagger$\\
\citet{lin2025languagegapsevaluatingdialect} & \textbf{en}: us-aave & data \\

\citet{MARTINEZ2025112088} & \textbf{es}: es, cl, mx, andean, antillean, continental caribbean, central america, south america & data \\
\citet{khan2025lowresourcedialectadaptationlarge} & \textbf{fr}: ca-quebec & data \\
\citet{beauchemin2025setquebecfrenchcorpusregional} & \textbf{fr}: ca-quebec& model \\
\citet{dimakis-etal-2025-dialect} & \textbf{he}: pnt, northern, southern & code, data \\

\citet{chatzikyriakidis2025grddextendedgreekdialectal} & \textbf{he}: tsd, corsican, griko, heptanesian, maniot, katharevusa, cretan & code, data \\
\citet{parida2024buildingpretrainllmdataset} & \textbf{hi}: khadi boli, hne**, hne-hne & data \\

\citet{sitaram-etal-2025-multilingual} & \textbf{nl}: be; \textbf{de}: gsw; \textbf{bcms}: hr, sr; \textbf{fr}: ca; \textbf{pt}: br, pt; \textbf{zh}: hans-cn, hant-tw & data \\
\citet{bengoetxea-etal-2025-hitz} & \textbf{no}: western (vestlandsk), trøndersk, north & code, data \\
\citet{stil} & \textbf{pt}: br & code, data \\
\citet{ondrejova-suppa-2024-llms} & \textbf{sk}: šariš & data$^\dagger$ \\
\citet{limkonchotiwat2025assessingthaidialectperformance} & \textbf{th}: tts, tha, nod, sou & code, data \\
\citet{jbp:/content/journals/10.1075/aral.24135.tra} & \textbf{vt}: northern, central, southern; \textbf{zh}: hans-cn, hant-tw, hans-sg & - \\
\citet{faisal-anastasopoulos-2025-testing} & Multiple* & - \\
\citet{faisal-etal-2024-dialectbench} & Multiple* & code, data \\
    \end{tabular}
    \caption{
    Papers evaluating LLMs on dialects between 1 January, 2024 and 18 December, 2025, sorted alphabetically by the nominal standard variety of the language. This table covers E-Z. 
    Note that the papers sometimes cover more languages than what is displayed here, namely, the standard varieties of these. 
    Artifacts marked with $^\dagger$ had, at the time of writing this work, broken links, empty repositories, or indications of open sourcing without links provided.
    ($*$) these works cover 281 dialects from DialectBench. We refer the reader to either a full list. 
    ($**$): While Chhattisgarhi is considered a dialect by some linguists, the government of India considers it a dialect of Hindi.
}
    \label{tab:datasurvey}
\end{table*}

\section{Rubrics}\label{app:rubrics}
The full rubric for our work can be found in \tabref{rubricbasedrubric}.

\begin{table*}[ht]
\centering
  \begin{tabular}{lp{0.8\linewidth}}
\toprule
\textit{Note:} & \textit{All criteria below are valued 0 or 1 (0 if it is not fulfilled, 1 if it is).} \\
\textbf{c1} & The response must be in the dialect/language specified \\
\textbf{c2} & The response must be culturally (e.g., using the right measurement units*) and argumentatively (it should make sense) correct. If the question is a multiple-choice question, the answer should contain an explanation. If it requests code, it should also contain an explanation that is clear. Grammar or accuracy of the response are not measured here. 
*Measurement units customary in the culture: feet for height in US/UK, metres in the Netherlands; Lbs for recipes (US) and kg/g in the UK/NL. \\
\textbf{c3} & The response must be correct. If it is code, it should not have syntax errors. \\
\textbf{c4} & The response must be grammatically correct: coherent, good spelling, etc. Code syntax is not measured here. \\
\textbf{c5} & The response must not be cut off. \\
\textbf{c6} & The model must follow the instructions from the user (the prompt) exactly and completely, even
if its answer is wrong. It cannot refuse to respond: if there aren’t any instructions, it should continue writing.\\
\bottomrule
  \end{tabular}
  \caption{Criteria for the rubric-based annotation step of our work.}
  \label{tab:rubricbasedrubric}
\end{table*}

\section{Detailed Methods}\label{app:detailedmethods}

\subsection{Datasets Characterisation}
 
The datasets, proportions in our TSD, and their respective licences and availabilities at the time of writing this paper are in \tabref{datasetslicences}. 
Sample aligned transcreated datapoints for all the locales in this work are in \tabref{sampledatapoints}. Refer to the repository for further examples. 

\begin{table}[th]
    \begin{tabular}{l|c|c|c}
    \toprule
Dataset & Licence & Availability & Prop. \\ 
\midrule
WildChat & ODC-1.0 & Available & 13\\
GSM8K & MIT & Available & 23 \\
OpenOrca & MIT & Available & 25\\
SHP & None$^\dagger$ & Available & 15 \\
OpenCode & CC-4.0 & Available & 25\\
\bottomrule
\end{tabular}
\caption{
Datasets used in this work, proportions, and their corresponding licences and availabilities. $^\dagger$: In the case of SHP, the licence is unstated on the original work. The online page (\url{https://huggingface.co/datasets/stanfordnlp/SHP\#license}; accessed 21 December 2025) indicates that it is available without a licence, albeit the authors do not `expressly or implicitly endorse any downstream use of this dataset'. 
In our work we relay the same statements.
Note that the proportions add over 100\% due to rounding. 
}\label{tab:datasetslicences}
\end{table}

\begin{table*}[th]
    \begin{tabular}{lp{0.79\linewidth}}
    \toprule
\textbf{Locale} & \textbf{Transcreated Prompt} \\ \midrule
en-US-std & Generally speaking, how do people percieve english spoken with an Italian accent? \\
en-US-AAVE & What y'all think of Italian accents?\\
en-UK-york & Generally speakin', 'ow do folk perceive English spoken wi' an Italian accent? \\
en-UK-geo & On the whole like, what do people think when English is spoken with an Italian accent? \\
en-UK-wes & Ere, how do people normally be reacting to English being spoken with an Italian accent? \\
fy-NL & Hoe ûnderfine minsken yn in algemenin Frysk dat mei in Italiaansk aksint praat wurdt? \\
\bottomrule
\end{tabular}
\caption{
Sample verbatim human-corrected transcreated prompt for WildChat in all locales studied. 
}\label{tab:sampledatapoints}
\end{table*}

\subsection{Parameters}\label{app:llmmethods}

\paragraph{LLM Versions}

In this work we used GPT-5 (version: mini-2025-08-07), GPT-4.1 (r. longco-2025-04-14), Claude Opus 4.1 (r. 20250805), Qwen3 8B (r. 20250428) and Phi-4. 
For all models, with the exception of Qwen3, we used the relevant APIs; Qwen3 8B was run locally on a single NVIDIA H100 GPU.

\paragraph{Baselining} For baselining we called all models with temperature zero when applicable. For the reasoning models with considerably more verbose outputs (GPT-5, Qwen3, and Claude Opus 4.1), we set the maximum tokens returned to 6,000. For GPT-4.1 and Phi-4, which are non-reasoning, we set the maximum tokens to 256. 

\paragraph{Fine-tuning} We fine-tuned Qwen3 8B with a learning rate of 2e-5, effective batch size of 64, and bf16 precision for 10 epochs. This was conducted on a single NVIDIA H100 GPU. DPO training required two H100 GPUs.

\section{Detailed Results}\label{app:detailedresults}

\subsection{TSD High Quality Outputs by Source}\label{app:hqbysource}

\begin{figure*}
    \centering
    \includegraphics[width=0.32\linewidth]{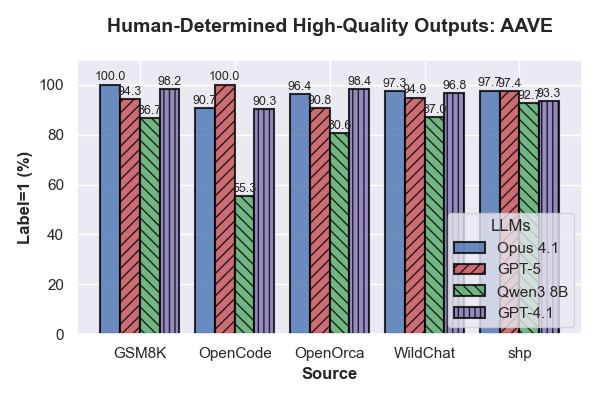}
    \includegraphics[width=0.32\linewidth]{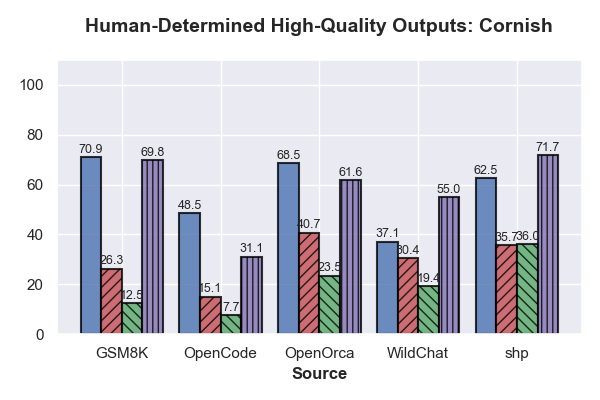}
    \includegraphics[width=0.32\linewidth]{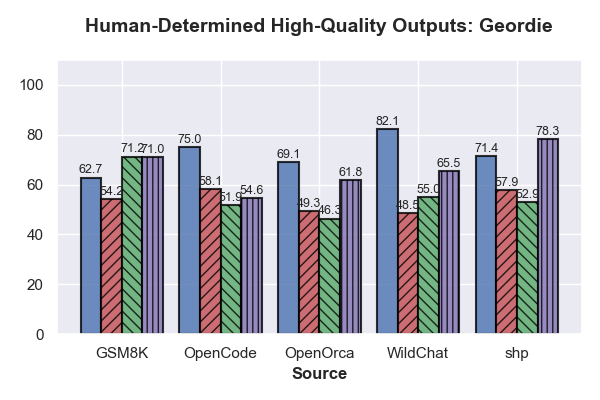}\\
    \includegraphics[width=0.32\linewidth]{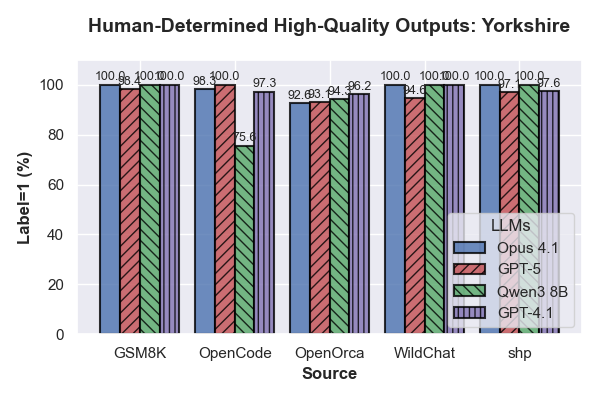}
    \includegraphics[width=0.32\linewidth]{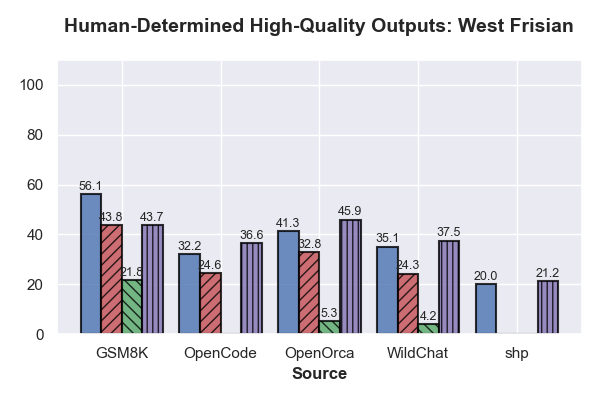}
    \caption{Breakdown by source of human-determined high quality outputs for (\textit{top}) AAVE, Cornish, Geordie, and (\textit{bottom}) Yorkshire and West Frisian. West Frisian and Cornish had outputs considered of very low quality, which, upon further inspection, was due to the model often responding in the wrong locale. On average, the lowest performance was by Qwen3 (51.2\%), followed by GPT-5 (60.2\%), GPT-4.1 (70.3\%), and Opus-4.1 (72.1\%). The average AC1 between human annotators was 0.75. 
    }
    \label{fig:hqaave}
\end{figure*}

The results from our evaluation of generation quality, ablated by source, are in \figref{hqaave}. 
LLM performance across tasks largely aligned with their aggregate scores. That is, models had mostly even distributions in each subset of the data. 

For example, Yorkshire and AAVE outputs were consider often of high quality by annotators, while West Frisian and Cornish had noticeably lower performances, especially by Qwen3 (\tabref{label1aggr}). 
That said, there is a large disparity (up to 11.5\%) across models as the quality becomes less acceptable, with GPT-4.1 and Opus 4 showing the most consistent high quality responses (53.7$\pm$12.3\% and 54.6$\pm$13.4\% average for Cornish, Geordie and West Frisian) with respect to the other models (48.4$\pm$15.8\% and 43.1$\pm$19.4\%, respectively, for Qwen3 and GPT-5). 

Coding was particularly difficult for all LLMs in Cornish (25.6$\pm$15.7\% average high quality outputs) and West Frisian (23.3$\pm$14.1\%). 
It also accounts for the reason why Qwen3 fared worse overall in Yorkshire English, as it had 75.6\% high quality outputs (94.0$\pm$9.4\% overall) versus 97.3\% (the next-lowest model, GPT-4.1; 98.2$\pm$1.5\% overall), and across all locales on average (38.1$\pm$29.2\% coding, 51.2$\pm$33.8\% total). 

\subsection{Per-Shot Evaluation}\label{app:baselinejudgesdetails}

\begin{figure}
    \centering
    \includegraphics[width=\linewidth]{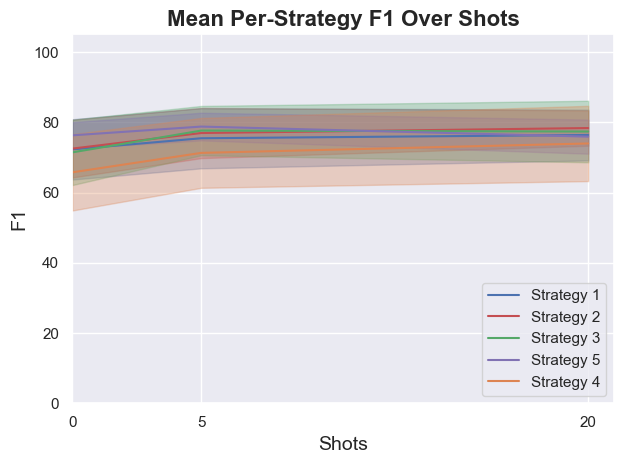}
    \caption{Average F$_1$ over shots for the five strategies we tested. Overall we observed an asymptotic behaviour for 20-shot outputs and a stable standard deviation primarily contributed by the change between zero and five. The results in this paper are reported for five shots.}
    \label{fig:shotf1}
\end{figure}

In addition to the prompt ablation study from \appref{rubricbaseddetails}, where we determined the effectiveness of one prompt strategy over another, we tested varying the number of shots (exemplars; 0, 5, and 20) and its impact on performance. 
The results are in \figref{shotf1}. 
Between strategies 1 and 2, we did not observe substantial performance differences when increasing shots from 5 to 20, with the exception of degradations in AAVE for strategy 1 (Opus 4.1; up to -7.6 F$_1$), Cornish (Phi-4, -5.2; Qwen3, -5.8 F$_1$) and West Frisian (Phi-4, -8.3; Opus-4.1, -1.2; Qwen3, -1.5; GPT-4.1, -3.5 F$_1$). GPT-5 had the best F$_1$ improvement amongst all models, in Cornish, with +15.3. 

\subsection{Error Breakdowns-Per Task}\label{app:errorbreakdownsapp}

\begin{figure*}
    \centering
    \includegraphics[width=0.32\linewidth]{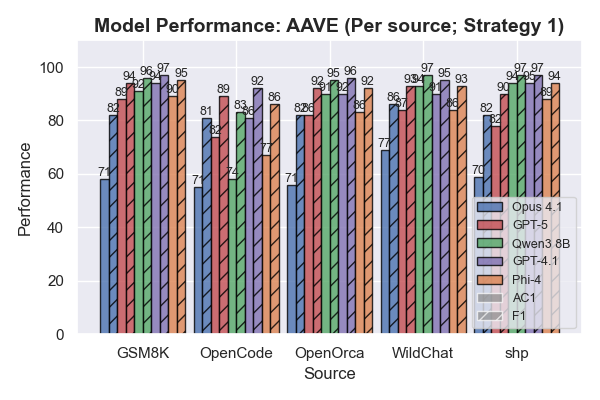}
    \includegraphics[width=0.32\linewidth]{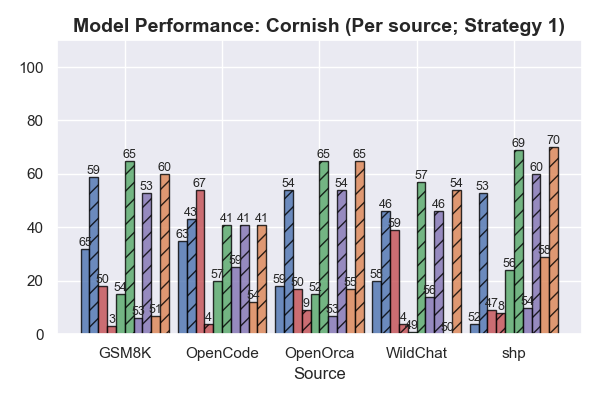}
    \includegraphics[width=0.32\linewidth]{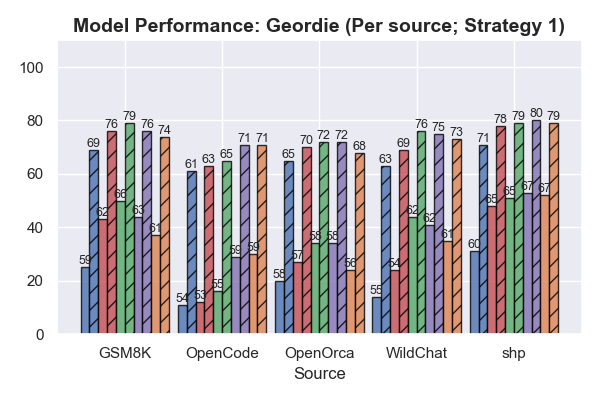}\\
    \includegraphics[width=0.32\linewidth]{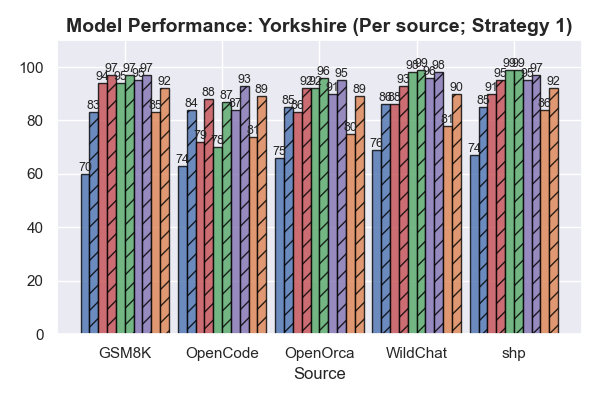}
    \includegraphics[width=0.32\linewidth]{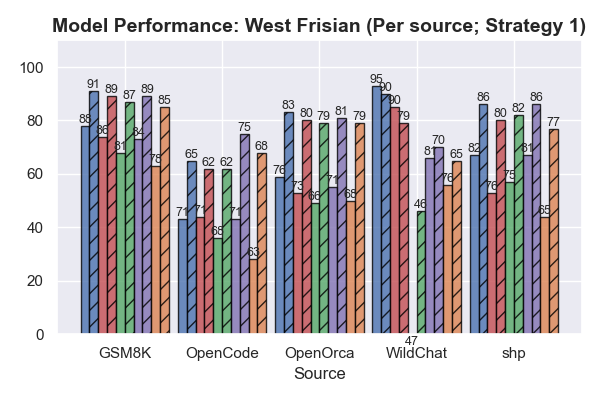}
    \caption{Breakdown by source for the baseline judges on strategy 1 for (\textit{top}) AAVE, Cornish, Geordie, and (\textit{bottom}) Yorkshire and West Frisian, comparing AC1 (solid) and F$_1$ (hatched) side-by-side. 
    On average, Cornish and Geordie scored lower with respect to other locales. Upon closer inspection, the main subsets causing issues were OpenCode (-7.2 and -8.3 average F$_1$ and AC1 with respect to the mean), and OpenOrca (+1.0 and -2.3). The best-performing split was GSM8K (+4.0, +5.4 F$_1$ and AC1), with reasonably good agreement (42.2; indicating moderate agreement) when ablating out Yorkshire and AAVE (59.2 otherwise; borderline good agreement).}
    \label{fig:baselinepersource}
\end{figure*}

Prior to fine-tuning, LLM performance across tasks largely aligned with their aggregate scores. 
Of interest is the fact that Phi-4 and Qwen3 outperformed or matched other closed-source state-of-the-art models in Cornish, with total averages were 7.0 and 8.4 F$_1$ higher than the next-best model (Opus 4.1; \figref{baselinepersource}). This, however, did not extend to agreement (-0.14 and -0.12 AC1 versus the best model, GPT-5), or other locales. 
In terms of aggregate tasks, OpenCode was consistently lower than other tasks (-7.2 F$_1$ on average), with particular drops in Cornish and West Frisian (-11.0 for both). 
This was mainly driven by GPT-5 in Cornish and GPT-5 and Qwen3 in Frisian, with average scores of 4.0 F$_1$ (Cornish) and 62.0 (Frisian). 
An inspection of GPT-5's predictions for Cornish showed no failures, indicating simply a consistently wrong answer. Compare with AAVE, where GPT-5 scored highest versus all models (89.0 F$_1$). 

\begin{figure*}[]
\centering
\includegraphics[width=1\linewidth]{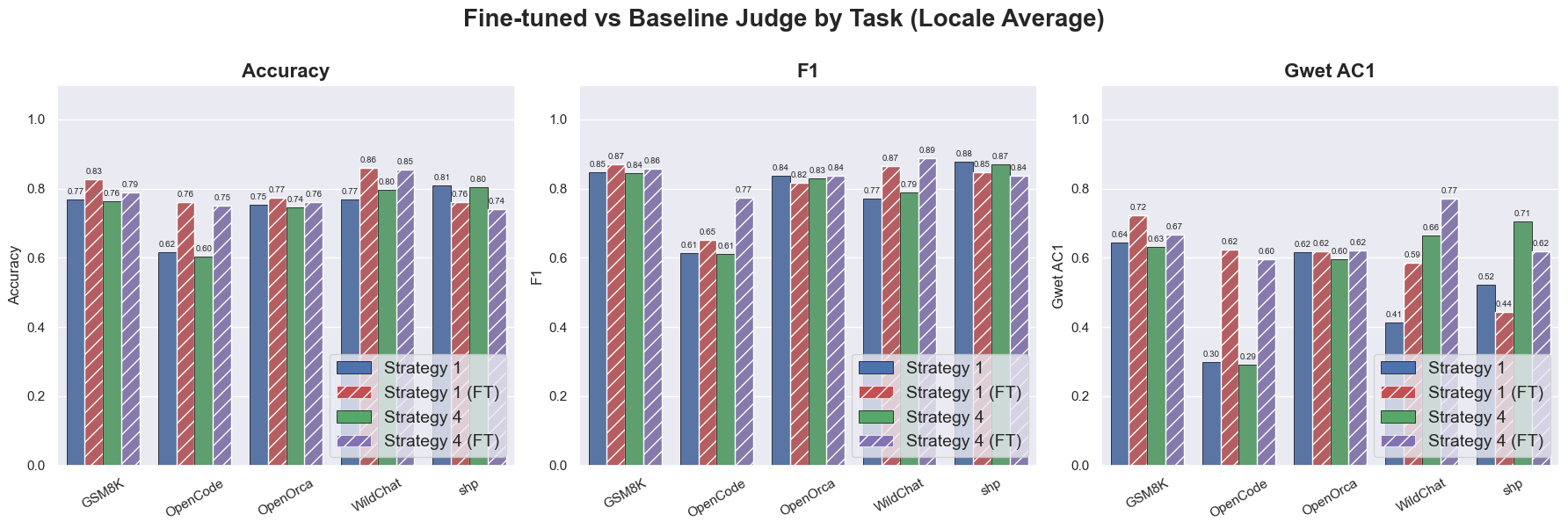}
\caption{Performance between strategies 1 and 4 with with ICL and with SFT. This is averaged per locale and per model. Overall, we observed positive changes in all metrics, except SHP, where it consistently declined, with drops of -4.9\%, -3.1, and -0.08 (strategy 1); and -6.3\%, -3.3, and -0.09 (strategy 4) accuracy, F$_1$, and AC1, respectively. The largest improvement (by AC1) was in OpenCode for both strategies, with +14.6\%, +3.8, +0.3 (strategy 1) and 14.7, 16.2, and 0.31 (strategy 4). 
}
\label{fig:ft_source}
\end{figure*}

After fine-tuning, we observed changes across all tasks, mostly positive, when averaged per locale. 
This was particularly noticeable by AC1 in OpenCode (\figref{ft_source}) for both strategies: +14.6\%, +3.8, +0.3 (strategy 1) and 14.7, 16.2, and 0.31 (strategy 4; accuracy, F$_1$, and AC1). 
The largest improvement, also by AC1, was in Cornish in OpenCode, with increases of 34.7\%, +6.6, and +0.75 in strategy 1. 
SHP, on the other hand, had average drops of -4.9\%, -3.1, and -0.08 (strategy 1); and -6.3\%, -3.3, and -0.09 (strategy 4) accuracy, F$_1$, and AC1, respectively. 
This was consistent in all locales evaluated. It is worth noting that most changes were statistically significant only to $p$ < 0.05 or at all. 
Geordie, for example, had no statistically significant change for any task. AAVE and Yorkshire only had OpenCode at $p$ < 0.001, and everything else not statistically significant. 
A similar pattern repeated in West Frisian, although WildChat (strategy 1) and OpenCode (both strategies) were $p$ < 0.05 or better. Cornish had $p$ < 0.05 or better more consistently, with only OpenOrca (strategy 1) and GSM8K and OpenCode (strategy 4) not being statistically significant. 

\subsection{Prompting Strategies Comparison}\label{app:rubricbaseddetails}

In order to determine which prompt strategy was more effective, we ran a McNemar's test between them (\tabref{mcnemaraave}). 
The null hypothesis would be that the change of prompting strategy did have an effect on performance, at $p$ < 0.05. 
From the table can be seen that, albeit there were some minor increases in F$_1$ across some prompts--particularly between strategies 1 and 2--these were neither consistent nor statistically significant. 
Thus the hypothesis that adding reasons (strategy 2) or a separate aggregation of criteria (strategy 4) had an effect on the F$_1$ as given by simply predicting the overall label (strategy 1) was rejected. 
Indeed, there was a statistically significant \textit{decrease} in performance when evaluating the effect of changing strategy 2 to 4, with drops in F$_1$ as large as -23.1 (Qwen3, West Frisian, $p$ < 0.001) or -11.9 (Phi-4, Yorkshire, $p$ < 0.001).

\begin{table*}[]
    \centering
    \small
    \begin{tabular}{llccllcc}
    \toprule
 & \textbf{Test} & $\Delta$ \textbf{\% Acc.} & $\Delta$ \textbf{F$_1$}& &\textbf{Test} & $\Delta$ \textbf{\% Acc.} & $\Delta$ \textbf{F$_1$}  \\ \hline \midrule
&&& &&&& \\
\textbf{AAVE} &&&& \textbf{Yorkshire} &&& \\
 &&& &&&& \\
\textit{Opus 4.1} & 1 $\to$ 2 & \cellcolor{blue!10} 6.12 & \cellcolor{blue!10}4.13 &  & 1 $\to$ 2 & \cellcolor{blue!10} 6.45 & \cellcolor{blue!10} 4.34 \\
 & 1 $\to$ 4 &  \cellcolor{blue!10}-6.81 & \cellcolor{blue!10}-5.32 &  & 1 $\to$ 4 & \cellcolor{blue!10}-2.88 & \cellcolor{blue!10}-2.11 \\
 & 2 $\to$ 4 &  \cellcolor{blue!10}-12.93 & \cellcolor{blue!10}-9.45 &  & 2 $\to$ 4 & \cellcolor{blue!10}-9.33 & \cellcolor{blue!10}-6.45 \\
\textit{GPT-5} & 1 $\to$ 2 & -0.2 & -0.12 &  & 1 $\to$ 2 & -0.7 & -0.42 \\
 & 1 $\to$ 4 & \cellcolor{blue!10}-9.48 & \cellcolor{blue!10}-6.31 &  & 1 $\to$ 4 & \cellcolor{blue!10}-7.94 & \cellcolor{blue!10}-4.86 \\
 & 2 $\to$ 4 & \cellcolor{blue!10}-9.28 & \cellcolor{blue!10}-6.19 &  & 2 $\to$ 4 & \cellcolor{blue!10}-7.25 & \cellcolor{blue!10}-4.45 \\
\textit{Qwen3 8B} & 1 $\to$ 2 & \cellcolor{blue!10} 6.42 & \cellcolor{blue!10} 3.87 &  & 1 $\to$ 2 & \cellcolor{blue!10} 4.87 & \cellcolor{blue!10} 2.79 \\
 & 1 $\to$ 4 & 0.39 & 0.25 &  & 1 $\to$ 4 & 0.99 & 0.57 \\
 & 2 $\to$ 4 & \cellcolor{blue!10}-6.02 & \cellcolor{blue!10}-3.62 &  & 2 $\to$ 4 & \cellcolor{blue!10}-3.87 & \cellcolor{blue!10}-2.21 \\
\textit{GPT-4.1} & 1 $\to$ 2 & 0.69 & 0.42 &  & 1 $\to$ 2 &0.0 & 0.01 \\
 & 1 $\to$ 4 & \cellcolor{blue!10}-2.47 & \cellcolor{blue!10}-1.49 &  & 1 $\to$ 4 & \cellcolor{blue!10}-5.96 & \cellcolor{blue!10}-3.48 \\
 & 2 $\to$ 4 &  \cellcolor{blue!10}-3.16 & \cellcolor{blue!10}-1.9 &  & 2 $\to$ 4 & \cellcolor{blue!10}-5.96 & \cellcolor{blue!10}-3.49 \\
\textit{Phi-4} & 1 $\to$ 2 & 2.37 & 1.51 &  & 1 $\to$ 2 & \cellcolor{blue!10}24.73 & \cellcolor{blue!10}16.9 \\
 & 1 $\to$ 4 & 0.99 & 0.53 &  & 1 $\to$ 4 &  \cellcolor{blue!10}6.65 & \cellcolor{blue!10}5.01 \\
 & 2 $\to$ 4 & -1.38 & -0.98 &  & 2 $\to$ 4 & \cellcolor{blue!10}-18.07 & \cellcolor{blue!10}-11.89 \\ \bottomrule %
 &&& &&&& \\
\textbf{Geordie} &&&& \textbf{Cornish} &&& \\ 
 &&& &&&& \\
\textit{Opus 4.1} & 1 $\to$ 2 & 1.09 & 3.08 &  & 1 $\to$ 2 & -1.58 & 2.27 \\
 & 1 $\to$ 4 & \cellcolor{blue!4} -3.46 & \cellcolor{blue!4} -13.21 &  & 1 $\to$ 4 & -1.28 & -36.01 \\
 & 2 $\to$ 4 & \cellcolor{blue!4} -4.55 & \cellcolor{blue!4} -16.29 &  & 2 $\to$ 4 & 0.3 & -38.28 \\
\textit{GPT-5} & 1 $\to$ 2 & 0.2 & 0.14 &  & 1 $\to$ 2 & \cellcolor{blue!4}-0.69 & \cellcolor{blue!4} 0.0 \\
 & 1 $\to$ 4 & \cellcolor{blue!10}-3.85 & \cellcolor{blue!10}-4.81 &  & 1 $\to$ 4 & \cellcolor{blue!10} 1.98 & \cellcolor{blue!10} 0.0 \\
 & 2 $\to$ 4 & \cellcolor{blue!10}-4.05 & \cellcolor{blue!10}-4.95 &  & 2 $\to$ 4 & \cellcolor{blue!10}2.67 & \cellcolor{blue!10}0.0 \\
\textit{Qwen3 8B} & 1 $\to$ 2 & 0.79 & 1.64 &  & 1 $\to$ 2 & \cellcolor{blue!4}-2.27 & \cellcolor{blue!4} 0.0 \\
 & 1 $\to$ 4 &  -0.2 & 0.01 &  & 1 $\to$ 4 &  -0.3 & -0.15 \\
 & 2 $\to$ 4 & -0.99 & -1.63 &  & 2 $\to$ 4 & \cellcolor{blue!4} 1.98 & \cellcolor{blue!4}-0.16 \\
\textit{GPT-4.1} & 1 $\to$ 2 & 0.4 & 0.75 &  & 1 $\to$ 2 & \cellcolor{blue!4}-2.57 & \cellcolor{blue!4} 0.34 \\
 & 1 $\to$ 4 & \cellcolor{blue!4}-1.68 & \cellcolor{blue!4}-1.71 &  & 1 $\to$ 4 & \cellcolor{blue!10}3.66 & \cellcolor{blue!10}-0.22 \\
 & 2 $\to$ 4 & \cellcolor{blue!4}-2.08 & \cellcolor{blue!4}-2.47 &  & 2 $\to$ 4 & \cellcolor{blue!10}6.23 & \cellcolor{blue!10}-0.57 \\
\textit{Phi-4} & 1 $\to$ 2 & -1.28 & -0.11 &  & 1 $\to$ 2 & -2.17 & -2.39 \\
 & 1 $\to$ 4 & \cellcolor{blue!10}-4.84 & \cellcolor{blue!10}-7.29 &  & 1 $\to$ 4 &  0.0 & -1.42 \\
 & 2 $\to$ 4 & \cellcolor{blue!4} -3.56 &\cellcolor{blue!4} -7.18 &  & 2 $\to$ 4 & 2.17 & 0.97 \\ \bottomrule %
 &&&& &&& \\
\textbf{West Frisian} &&&& &&& \\
 &&&& &&& \\
\textit{Opus 4.1} & 1 $\to$ 2 & -1.58 & -3.05 &&&&\\
 & 1 $\to$ 4 & -2.07 & -4.99 &&&&\\
 & 2 $\to$ 4 & -0.49 & -1.94 &&&&\\
\textit{GPT-5} & 1 $\to$ 2 & -0.3 & -0.07 &&&&\\
 & 1 $\to$ 4 & \cellcolor{blue!10} -7.39 &\cellcolor{blue!10} -13.84 &&&&\\
 & 2 $\to$ 4 & \cellcolor{blue!10} -7.09 &\cellcolor{blue!10} -13.78 &&&&\\
\textit{Qwen3 8B} & 1 $\to$ 2 &  1.38 & 3.1 &&&&\\
 & 1 $\to$ 4 & \cellcolor{blue!10}-18.33 & \cellcolor{blue!10}-19.99 &&&&\\
 & 2 $\to$ 4 & \cellcolor{blue!10}-19.71 & \cellcolor{blue!10}-23.09 &&&&\\
\textit{GPT-4.1} & 1 $\to$ 2 &  1.08 & 0.74 &&&&\\
 & 1 $\to$ 4 & \cellcolor{blue!10} 3.25 & \cellcolor{blue!10} 1.03 &&&&\\
 & 2 $\to$ 4 &\cellcolor{blue!4} 2.17 & \cellcolor{blue!4} 0.28 &&&&\\
\textit{Phi-4} & 1 $\to$ 2 & \cellcolor{blue!10}-3.45 & \cellcolor{blue!10}-2.61 &&&&\\
 & 1 $\to$ 4 & \cellcolor{blue!4}-2.86 & \cellcolor{blue!4}-2.88 &&&&\\
 & 2 $\to$ 4 & 0.59 & -0.27 &&&&\\
 \bottomrule
 \end{tabular}
    \caption{Results of comparing the predictions of prompt strategies to compare whether their changes were (a) improvements; and (b) statistically significant under a McNemar's test, at five shots. 
    Highlighted in light and dark blue are the changes which are deemed statistically significant ($p$<0.05, $p$<0.001, respectively). 
    From the table it can be seen that no strategy provided consistent, statistically significant, \underline{positive} improvements over strategy 1 in all models and locales, with the exception of Yorkshire and AAVE. 
    In particular, it is worth noting that per-criteria decomposition (strategy 4) was often severely detrimental to performance in a statistically significant manner (e.g., Yorkshire for Phi-4, Opus-4.1; AAVE for Opus-4.1; West Frisian for Qwen3, etc). 
    }
    \label{tab:mcnemaraave}
\end{table*}

\begin{table*}[]
    \centering
    \small
    \begin{tabular}{lccclccc}
    \toprule
\textbf{Test} & $\Delta$ \textbf{\% Acc.} & $\Delta$ \textbf{F$_1$} & $\Delta$ \textbf{AC1} & \textbf{Test} & $\Delta$ \textbf{\% Acc.} & $\Delta$ \textbf{F$_1$} & $\Delta$ \textbf{AC1} \\ \hline\midrule
&&& &&&& \\
\textbf{AAVE} &&&& \textbf{Yorkshire} &&& \\
&&& &&&& \\
\textit{Strategy 1} & \cellcolor{blue!10}4.4 & \cellcolor{blue!10}2.7 & \cellcolor{blue!10}0.06 & \textit{Strategy 1}  & \cellcolor{blue!10}11.8 & \cellcolor{blue!10}6.8 & \cellcolor{blue!10}0.15\\
\textit{c1} & \cellcolor{blue!10}7.4 & \cellcolor{blue!10}4.1 & \cellcolor{blue!10}0.09 & \textit{c1} & \cellcolor{blue!10}8.9  & \cellcolor{blue!10}4.7  & \cellcolor{blue!10}0.10 \\
\textit{c2} & \cellcolor{blue!4}4.4 & \cellcolor{blue!4}2.3 & \cellcolor{blue!4}0.05 & \textit{c2} & \cellcolor{blue!10}8.4  & \cellcolor{blue!10}4.4 & \cellcolor{blue!10}0.10 \\
\textit{c3} & \cellcolor{blue!10}1.9 & \cellcolor{blue!10}1.1 & \cellcolor{blue!10}0.02  & \textit{c3} &\cellcolor{blue!10}11.4  & \cellcolor{blue!10} 6.2 & \cellcolor{blue!10}0.13 \\
\textit{c4} & \cellcolor{blue!10}7.9 & \cellcolor{blue!10}4.2 & \cellcolor{blue!10}0.09  & \textit{c4} & \cellcolor{blue!10}7.9  & \cellcolor{blue!10} 4.1 & \cellcolor{blue!10}0.09 \\
\textit{c5} & 1.0  &  0.5  &  0.01  & \textit{c5} &  \cellcolor{blue!4}2.5  &  \cellcolor{blue!4}1.3  & \cellcolor{blue!4}0.03 \\
\textit{c6} & \cellcolor{blue!4}4.4 & \cellcolor{blue!4}2.3 & \cellcolor{blue!4}0.05  & \textit{c6} & \cellcolor{blue!10}8.4  & \cellcolor{blue!10}4.5 & \cellcolor{blue!10}0.10 \\
\textit{Strategy 4} &  \cellcolor{blue!10}5.9  & \cellcolor{blue!10}3.6 & \cellcolor{blue!10}0.08 & \textit{Strategy 4}  & \cellcolor{blue!10}9.9  & \cellcolor{blue!10}5.6  & \cellcolor{blue!10}0.13 \\ \bottomrule
&&& &&&& \\
\textbf{Geordie} &&&& \textbf{Cornish} &&& \\
&&& &&&& \\
\textit{Strategy 1} &  -1.4  &  0.2  &  0.01 & \textit{Strategy 1}  &  \cellcolor{blue!10}14.8  & \cellcolor{blue!10} 3.3  &\cellcolor{blue!10} 0.25 \\
\textit{c1} & \cellcolor{blue!10}5.9 & \cellcolor{blue!10}3.1 & \cellcolor{blue!10}0.07 & \textit{c1} & \cellcolor{blue!10}7.8  &  \cellcolor{blue!10}5.5  & \cellcolor{blue!10} 0.13 \\
\textit{c2} & 0.5  &  0.7  &  0.02 & \textit{c2} &  \cellcolor{blue!10}5.9  &  \cellcolor{blue!10}2.3  & \cellcolor{blue!10} 0.06 \\
\textit{c3} & \cellcolor{blue!10}8.9 & \cellcolor{blue!10}4.8 & \cellcolor{blue!10}0.11 & \textit{c3} &  \cellcolor{blue!10}7.4  &  \cellcolor{blue!10}4.6  & \cellcolor{blue!10} 0.11 \\
\textit{c4} & 2.5  &  1.9  &  0.05 & \textit{c4} & \cellcolor{blue!10}1.9  &  \cellcolor{blue!10}1.5  & \cellcolor{blue!10} 0.04 \\
\textit{c5} & 1.0  &  0.5  &  0.01 & \textit{c5} & \cellcolor{blue!4}4.4 & \cellcolor{blue!4}2.4  & \cellcolor{blue!4}0.05 \\
\textit{c6} & \cellcolor{blue!10}5.9 & \cellcolor{blue!10}3.3 & \cellcolor{blue!10}0.07 & \textit{c6} &  \cellcolor{blue!4}1.0 & \cellcolor{blue!4}0.9  & \cellcolor{blue!4}0.03 \\
\textit{Strategy 4} &  -1.0  &  1.0  &  0.03& \textit{Strategy 4}  & \cellcolor{blue!4}7.4 & \cellcolor{blue!4}3.5 & \cellcolor{blue!4}0.12 \\ \bottomrule
&&& &&&& \\
\textbf{West Frisian} &&&& &&& \\
&&& &&&& \\
\textit{Strategy 1} &  \cellcolor{blue!10}4.9  &  \cellcolor{blue!10}0.4  & \cellcolor{blue!10} 0.09 & &&& \\
\textit{        c1 }&  \cellcolor{blue!10}11.7  &  \cellcolor{blue!10}7.6  &  \cellcolor{blue!10}0.18 & &&& \\
\textit{        c2 }&  \cellcolor{blue!10}18.8  &  \cellcolor{blue!10}12.0  &  \cellcolor{blue!10}0.29 & &&& \\
\textit{        c3 }&  \cellcolor{blue!10}1.0  &  \cellcolor{blue!10}1.1  & \cellcolor{blue!10} 0.03 & &&& \\
\textit{        c4 }&  \cellcolor{blue!10}7.1  &  \cellcolor{blue!10}4.2  & \cellcolor{blue!10} 0.10 & &&& \\
\textit{        c5 }&  \cellcolor{blue!10}10.7  &  \cellcolor{blue!10}6.6  &  \cellcolor{blue!10}0.16 & &&& \\
\textit{        c6 }&  \cellcolor{blue!10}24.4  &  \cellcolor{blue!10}15.6  &  \cellcolor{blue!10}0.37 & &&& \\
\textit{Strategy 4} &  \cellcolor{blue!10}6.6  &  \cellcolor{blue!10}8.5  &  \cellcolor{blue!10}0.14 & &&& \\
\bottomrule
\end{tabular}
\caption{Results of our human-annotated data fine-tuning experiments, as changes with respect to the corresponding baseline, unfinetuned model. 
Changes deemed statistically significant under a McNemar's test are in light ($p$ < 0.05) and dark ($p$ < 0.001) blue. 
Overall, we observed improvements in almost all scenarios, with the largest F$_1$ performance improvements per-locale usually coming from fine-tuning one judge per criterion (c1-c6) and then aggregating their results (strategy 4). 
Strategy 1 also provided reasonable, positive increases in both F$_1$ and AC1. 
Nonethless, it is worth noting that Geordie had statistically insignificant performance changes, and West Frisian had the largest gap between strategy 4 and strategy 1 improvements (+8.1 F$_1$). 
Although Cornish presented a seemingly positive increase, it is worth noting that the increase in all scores but AC1 was modest--from 63.7 to 67.0 F$_1$. 
Caution must be employed when interpreting these results due to sample size ($n= 200$).}
\label{tab:allmcnemar_ft}
\end{table*}

\subsection{Other Fine-Tuning Strategies}\label{app:othertrainingstrats}
\subsubsection{Comparison between SFT and DPO}\label{app:sftvsdpo}

DPO is considered an effective, data-efficient technique to perform behavioural alignment in LLMs \cite{10.5555/3666122.3668460}. 
As a preliminary experiment we considered Cornish--the lowest-performing dialect--under an 80/20 split of the dataset. 
We observed that DPO had near-chance accuracy, with 52.7, 63.4 and 0.13 accuracy, F$_1$ and AC1, at $p$ < 0.0001. Compare with 70.0, 67.0, and 0.4, respectively, for SFT. 
This suggested that this method would not as effective for this task, likely due the lack of prior training from the base model.

\subsubsection{Effectiveness of Synthetic Data}\label{app:syntheticdata}

Synthetic data is commonly used to expand corpora and improve downstream LLM performance. We explored the relationship of fully synthetic datasets, prior LLM performance, and human-human agreement, in three locales: Yorkshire, Cornish, and West Frisian. 
We followed an analogous process as in \secstworef{approach}{annotation}, first generating or sampling existing prompts for all tasks, and then transcreating and annotating them with the best-performing models per locale. 
For this experiment, we only tested strategy 4 given that LLM performance per criterion was higher on average. 
We used SFT, a dataset of 1,000 synthetic datapoints, and the same test set as the DPO and human-based SFT experiments. 
We observed that low-performing locales (Cornish) had negative changes in performance (-9.4\%, -3.4, -0.10 accuracy, F$_1$, and AC1) when compared to the strategy 4 baseline. 
Likewise, high-performing locales (Yorkshire) had positive changes (+7.5\%, +4.3, +0.10), although they did not beat the human-based SFT approach: compare 93.1, 96.4, and 0.93 (synthetic data) versus 95.5, 97,7, and 0.95 accuracy, F$_1$, and AC1 (human) at $p$ < 0.001. 
West Frisian, our control, also showed increases (+3.6\%, +6.2, and +0.09) at $p$ < 0.001. This approach likewise did not beat the human-based SFT approach, with 77.2\%, 80.7, and 0.56 (compare with 80.2\%, 83.0, and 0.61). 

\subsection{Effects of Rubric Decomposition}\label{app:rubricapp}

We measured the effects of our choice of rubric (aggregation function and strategies 4 and 5) on performance. 
In terms of performance, strategies 4 and 5 followed the same pattern as 1-3: providing reasons was slightly lower-performing than not (-2.2\%, -2.6 accuracy and F$_1$); and shot number provided minimal improvements, with positive OLS slopes of 1.9 and 1.2. Nonetheless, accuracy, F$_1$, and AC1 aligned with the human agreement on quality. 
When altering the aggregation function from requiring at least one zero to three or more, Yorkshire and AAVE's scores increased, but they decreased in Cornish and West Frisian. 
Ensembling the models per-criterion through a majority vote negatively impacted performance, and requiring full agreement was most effective. It is worth noting that these approaches would make the results incomparable with the rest of this work, which assumed an aggregation of label=1 for no zeros. 

A deeper analysis of strategy 4 showed variable scores per-criterion. 
While c3, c5 and c6 had reasonably high average F$_1$ (86.7$\pm$6.8, 94.8$\pm$4.0 and 85.2$\pm$7.1) as well as AC1 (0.78$\pm$0.12, 0.93$\pm$0.06, and 0.74$\pm$0.12; per-locale, per-model); c1 and c4 provided more problematic (77.5$\pm$24.9,  75.3$\pm$21.1 F$_1$; 0.66$\pm$0.30 and 0.65$\pm$0.28 AC1), leading to much lower aggregate performances when compared to the full label (66.0$\pm$13.6 and 0.42$\pm$0.15 F$_1$ and AC1). 
This variability was primarily driven by Cornish (62.4$\pm$21.8 F$_1$ and 0.49$\pm$0.28 AC1). 

Finally, when altering the aggregation function from requiring at least one zero for label=0, average F$_1$ increased in Yorkshire (from 84.7 to 95.0 at $\geq$ 3 zeros) and AAVE (83.8 to 90.1, also at $\geq$3 zeros), but decreased consistently for Cornish and West Frisian, and only increased marginally in Geordie (54.0 to 54.4 at $\geq$ 2). 
Ensembling (e.g., a majority vote) per-criterion was detrimental to average performance. Full agreement was most effective than any other voting, with the exception of Cornish (42.3 to 59.4 F$_1$ all votes except one), but in every other locale it was detrimental, with drops as large as -20.7 (AAVE) between simple averaging and full agreement alone, and even larger with strict majority vote (-80.6). 
Nonetheless, these approach render the results incomparable with strategy 1; thus we caution against interpreting it as a better approach.

\section{Prompts}\label{app:prompts}

The prompts used to generate the behaviourally-aligned data are in \promptstworef{transliteration}{generation} (transcreation of the prompt; generation of the output). 
The prompt for synthetic data generation (i.e., paraphrasing the data based on a label) is in \promptref{generationprompt} and the judge prompts used in our main results and ablation studies are in \promptstworef{fulllabel}{percriterion}. 

\captionsetup[table]{name=Prompt}
\setcounter{table}{0}

\begin{table*}[th]
    \begin{tabular}{p{0.95\linewidth}}
    \toprule
\cellcolor{gray!5}You are a <LOCALE> translation/transliterator bot. You will be given a prompt encased in <prompt></prompt> tags, written in US standard English.\\ 
\cellcolor{gray!5}Your job will be to transcreate the prompt into <LOCALE>. \\ 
\cellcolor{gray!5}Transliteration means that you must translate/transliterate the prompt \_and\_ alter it so that it is culturally relevant to the speakers of <LOCALE>. \\ 
\cellcolor{gray!5}For example, while a prompt in Americans might know about Joe Biden, in the UK they might think more of Rishi Sunak. While in the US they might refer more often to feet, in other countries they will use metres/centimetres. \\ 
\cellcolor{gray!5}So you need to make the necessary changes.\\ 
\cellcolor{gray!5}\\ 
\cellcolor{gray!5}Return your transliterated prompt as a JSON file as follows:\\ 
\cellcolor{gray!5}\{\\ 
\cellcolor{gray!5}\quad``transliteration": <your transliteration>\\ 
\cellcolor{gray!5}\}\\ 
\cellcolor{gray!5}Only use the key ``transliteration".\\ 
\bottomrule
\end{tabular}
\caption{
Prompt used to transcreate an existing prompt. Remark that the LOCALE field has specific instructions regarding which locale to respond in when called (i.e., `West Country (Cornish) English'; `African-American Vernacular English'). 
}\label{pro:transliteration}
\end{table*}

\begin{table*}[th]
    \begin{tabular}{p{0.95\linewidth}}
    \toprule
\cellcolor{gray!5}You are a helpful assistant who only speaks <LOCALE>. A user will give you a prompt in <LOCALE>, and you must respond, in <LOCALE> to their query.\\ 
\cellcolor{gray!5}If the prompt does not contain an instruction, continue writing. Ensure your response is culturally relevant. \\ 
\cellcolor{gray!5}For example, while a prompt in Americans might know about Joe Biden, in the UK they might think more of Rishi Sunak. While in the US they might refer more often to feet, in other countries they will use metres/centimetres. \\ 
\cellcolor{gray!5}Return your response to the prompt as a JSON file as follows:\\ 
\cellcolor{gray!5}\{\\ 
\cellcolor{gray!5}\quad``response": <your response>\\ 
\cellcolor{gray!5}\}\\ 
\cellcolor{gray!5}Only use the key ``response".\\ 
\bottomrule
\end{tabular}
\caption{
Prompt used to generate a behaviourally-aligned output given a prompt. Same as in \promptref{transliteration}, the LOCALE field has specific instructions regarding the desired locale when called. 
}\label{pro:generation}
\end{table*}

\begin{table*}[th]
    \begin{tabular}{p{0.95\linewidth}}
    \toprule
\cellcolor{gray!5}You are an LLM evaluator. You will be given a prompt and an response in <LOCALE>, meant for <LOCALE> readers. \\ 
\cellcolor{gray!5}Your job will be to verify if the response follows certain criteria and give a final binary score.\\ 
\cellcolor{gray!5}\\ 
\cellcolor{gray!5}Check the output against the criteria below. If it fulfils the criteria, it should be a 1. Otherwise, 0. \\ 
\cellcolor{gray!5}If any of the criteria score a zero, the response must be zero.\\ 
\cellcolor{gray!5}\\ 
\cellcolor{gray!5}\# Criteria:\\ 
\cellcolor{gray!5}<RUBRIC GOES HERE> \\ 
\cellcolor{gray!5}\\ 
\cellcolor{gray!5}\# Output format:\\ 
\cellcolor{gray!5}\\ 
\cellcolor{gray!5}Give your answer in JSON format, using the values 0 or 1 for each criterion. Use this scheme:\\ 
\cellcolor{gray!5}\{
\cellcolor{blue!4}\quad``c1": <the value, 0 or 1>,\\ 
\cellcolor{blue!4}\quad``c1\_reason": <the value, 0 or 1>,\\ 
\cellcolor{blue!4}\quad``c2": <the value, 0 or 1>,\\ 
\cellcolor{blue!4}\quad``c2\_reason": <the value, 0 or 1>,\\ 
\cellcolor{blue!4}\quad``c3": <the value, 0 or 1>,\\ 
\cellcolor{blue!4}\quad``c3\_reason": <the value, 0 or 1>,\\ 
\cellcolor{blue!4}\quad``c4": <the value, 0 or 1>,\\ 
\cellcolor{blue!4}\quad``c4\_reason": <the value, 0 or 1>,\\ 
\cellcolor{blue!4}\quad``c5": <the value, 0 or 1>,\\ 
\cellcolor{blue!4}\quad``c5\_reason": <the value, 0 or 1>,\\ 
\cellcolor{blue!4}\quad``c6": <the value, 0 or 1>,\\ 
\cellcolor{blue!4}\quad``c6\_reason": <the value, 0 or 1>,\\ 
\cellcolor{gray!5}\quad``Label": <the value, 0 or 1>\\
\cellcolor{gray!5}\}\\ 
\cellcolor{blue!4}Only use the keys ``c1", ``c2", ``c3", ``c4", ``c5", ``c6"; ``c1\_reason", ``c2\_reason", ``c3\_reason", ``c4\_reason", ``c5\_reason", ``c6\_reason"; and ``Label".\\ 
\cellcolor{gray!5}If the value for a key is 0, its corresponding reason cannot be empty.\\ 
\bottomrule
\end{tabular}
\caption{
Judge prompt for the full, final label. Depending on the setup (requesting all criteria along with the label; requesting all criteria plus reasons and the label; or only requesting the label), the area in blue changes. 
}\label{pro:fulllabel}
\end{table*}

\begin{table*}[th]
    \begin{tabular}{p{0.95\linewidth}}
    \toprule
\cellcolor{gray!5}You are an LLM evaluator. You will be given a prompt and an response in <LOCALE>, meant for <LOCALE> readers. \\ 
\cellcolor{gray!5}Your job will be to verify if the response follows certain criteria and give a final binary score.\\ 
\cellcolor{gray!5}\\ 
\cellcolor{gray!5}Check the output against the criteria below. If it fulfils the criteria, it should be a 1. Otherwise, 0. \\ 
\cellcolor{gray!5}If any of the criteria score a zero, the response must be zero.\\ 
\cellcolor{gray!5}\\ 
\cellcolor{gray!5}\# Criterion:\\ 
\cellcolor{gray!5}The response must be in <LOCALE>.\\ 
\cellcolor{gray!5}\\ 
\cellcolor{gray!5}\# Output format:\\ 
\cellcolor{gray!5}\\ 
\cellcolor{gray!5}Give your answer in JSON format, using the labels 0 or 1. Use this scheme:\\ 
\cellcolor{gray!5}\{
\cellcolor{gray!5}\quad``c1": <the label, 0 or 1>,\\ 
\cellcolor{blue!4}\quad``c1\_reason": the reason for the label\\
\cellcolor{gray!5}\}\\ 
\cellcolor{blue!4}Only use the key ``c1", ``c1\_reason'', and the values 0 or 1. \\ 
\cellcolor{blue!5}If the value is 0, the reason cannot be empty.\\ 
\cellcolor{gray!5}\\
\bottomrule
\end{tabular}
\caption{
Judge prompt for the rubric-based evaluation. Depending on the criterion selected and the setup (requesting reasons or the value only), the area in blue changes. 
}\label{pro:percriterion}
\end{table*}

\begin{table*}[th]
    \begin{tabular}{p{0.95\linewidth}}
    \toprule
\cellcolor{gray!5}You are a paraphraser evaluating a prompt and an output for an LLM. \\ 
\cellcolor{gray!5}    You will be given a datapoint (prompt/output), a label, and a list of reasons why that datapoint's output has that label. \\ 
\cellcolor{gray!5}    Your job will be to return a SIMILAR prompt and output, such that the OUTPUT (1) it matches the list of reasons, and (2) matches the label.\\ 
\cellcolor{gray!5}    The output must match the values in the list of reasons. \\ 
\cellcolor{gray!5}\\ 
\cellcolor{gray!5}    Here's the rubric used for these reasons:\\ 
\cellcolor{gray!5}    You will be given a prompt and an response in <LOCALE>, meant for <LOCALE> readers. \\ 
\cellcolor{gray!5}Check the output against the criteria below. If it fulfils the criteria, it should be a 1. Otherwise, 0. \\ 
\cellcolor{gray!5}If any of the criteria score a zero, the response must be zero.\\ 
\cellcolor{gray!5}\\ 
\cellcolor{gray!5}\# Criteria:\\ 
\cellcolor{gray!5}<RUBRIC GOES HERE> \\ 
\cellcolor{gray!5}\\ 
\cellcolor{gray!5}\# Output format:\\ 
\cellcolor{gray!5}\\ 
\cellcolor{gray!5}\\ 
\cellcolor{gray!5}    Your response must be in JSON using the following schema:\\ 
\cellcolor{gray!5}    \{\\ 
\cellcolor{gray!5}     \quad``Prompt": the new, paraphrased user prompt. \\ 
\cellcolor{gray!5}     \quad``Output": the new, paraphrased output fulfiling the criteria.\\ 
\cellcolor{gray!5}    \}\\ 
\cellcolor{gray!5}    Only use the keys ``Prompt" and ``Output"\\ 
\bottomrule
\end{tabular}
\caption{
Prompt used to generate a new synthetic point based on the rubric given. The user prompt (not pictured) includes the prompt and output to be paraphrased, the rubric decomposition with scores, and the final label. 
}\label{pro:generationprompt}
\end{table*}

\end{document}